\documentclass[lettersize,journal]{IEEEtran} 
\usepackage{bbm}
\usepackage{amsmath,amssymb,amsfonts}
\usepackage{algorithm}
\usepackage{cite}
\usepackage{array}
\usepackage{caption}
\usepackage{textcomp}
\usepackage{stfloats}
\usepackage{url}
\usepackage{verbatim}
\usepackage{graphicx,color}
\usepackage[dvipsnames]{xcolor}
\usepackage{hyperref}
\usepackage{subcaption}
\usepackage{algpseudocode}
\usepackage[nolist]{acronym}
\usepackage{cleveref}
\usepackage{tikz}
\usepackage{tabto}
\usepackage{booktabs}
\usepackage{adjustbox} 
\acrodef{AI}{Artificial Intelligence}
\acrodef{ML}{Machine Learning}
\acrodef{DL}{Deep Learning}
\acrodef{NN}{Neural Network}
\acrodef{RIS}{Reconfigurable Intelligent Surfaces}
\acrodef{KKT}{Karush-Kuhn-Tucker}
\acrodef{MIMO}{Multiple-Input Multiple-Output}
\acrodef{AWGN}{Additive White Gaussian Noise}
\acrodef{SNR}{Signal to Noise Ratio}
\acrodef{ADMM}{Alternating Direction Method of Multipliers}
\acrodef{MSE}{Mean Squared Error}
\acrodef{SVD}{Singular Value Decomposition}
\acrodef{RL}{Reinforcement Learning}
\acrodef{RIS}{Reconfigurable Intelligent Surfaces}
\acrodef{RR}{Relative Representation}
\acrodef{6G}{Sixth-Generation}
\acrodef{MIMO}{Multiple-Input Multiple-Output}
\acrodef{SISO}{Single-Input Single-Output}
\acrodef{JSCC}{Joint Source-Channel Coding}
\acrodef{DJSCC}{Deep JSCC}
\acrodef{OFDM}{Orthogonal Frequency-Division Multiplexing}
\acrodef{MDP}{Markov Decision Process}
\acrodefplural{MDP}{Markov Decision Processes}
\acrodef{VQ-VAE}{Vector Quantized Variational Autoencoder}
\acrodef{DNN}{Deep Neural Network}
\acrodef{PGD}{Projected Gradient Descent}
\acrodef{IoT}{Internet of Things}
\acrodef{SemComs}{Semantic Communications}
\acrodef{MMSE}{Minimum Mean Squared Error}
\acrodef{FLOP}{Floating Point Operation}
\acrodef{GELU}{Gaussian Error Linear Unit}

\def\data{{\mathbf x}}
\def\latent{{\mathbf z}}
\def\latentr{{\mathbf s}}
\def \noise{{\mathbf w}} 
\def \g{{\mathbf g}} 
\def \f{{\mathbf f}} 
\def \action{{\mathbf y}} 
\def \v{{\mathbf v}} 

\def\dspace{{\mathcal X}} 
\newcommand{\dataset}{\mathbf{X}} 
\def\cY{{\mathcal Y}} 
\def\bR{{\mathbb R}}
\def\bC{{\mathbb C}}
\def\bE{{\mathbb E}}

\def\enc{{E}}
\def\dec{{D}}
\def \ris{{\mathbf \phi}}
\def \rism{{\text{diag}\left(\ris\right)}} 
\renewcommand{\H}{\mathbf{H}}

\def\sw{{\theta}} 
\def\tw{{\gamma}} 

\def\slatent{\latent_\sw} 
\def\slatenti{\slatent^{(i)}}
\def\tlatent{\latent_\tw}
\def\tlatenti{\tlatent^{(i)}}
\def\slatentr{{\latentr_\sw}}
\def\slatentri{{\slatentr^{(i)}}}
\def\tlatentr{{\latentr_\tw}}
\def\tlatentri{{\tlatentr^{(i)}}}

\def\ldata{{\mathbf{Z}}}
\def\ldatareal{{\mathbf{S}}}
\def\sdata{\ldata_\sw} 
\def\sdatareal{\ldatareal_\sw}
\def\tdata{\ldata_\tw} 
\def\tdatareal{\ldatareal_\tw}

\def\tenc{{\enc_\tw}} 
\def\senc{{\enc_\sw}}
\def\tdec{{\dec_\tw}} 
\def\sdec{{\dec_\sw}}

\def\sdim{{N_\sw}} 
\def\tdim{{N_\tw}}
\def\sspace{\bR^{\sdim}}
\def\tspace{\bR^{\tdim}}
\def\sdimC{{\frac{\sdim}{2}}} 
\def\tdimC{{\frac{\tdim}{2}}}
\def\sspaceC{\bC^{\sdimC}}
\def\tspaceC{\bC^{\tdimC}}

\def\F{{\mathbf F}}
\def\G{{\mathbf G}}
\def\V{{\mathbf V}}
\def\U{{\mathbf U}}
\def\D{{\mathbf D}}
\def\R{{\mathbf R}}
\def\S{{\mathbf S}}
\def\T{{\mathbf T}}
\def\I{{\mathbf I}}

\newcommand{\abs}[1]{\left|#1\right|}

\newcommand{\norm}[1]{\left\|#1\right\|}
\newcommand{\parenthesis}[1]{\left(#1\right)}
\newcommand{\squareb}[1]{\left[#1\right]}
\newcommand{\curlyb}[1]{\left\{#1\right\}}

\DeclareMathOperator*{\argmin}{arg\,min}

\newcommand{\quotes}[1]{ `#1'}
\crefname{equation}{}{}
\crefname{figure}{Figure}{Figures}

\usetikzlibrary{positioning}
\usetikzlibrary{shapes.multipart}
\usetikzlibrary{shapes.geometric}
\usetikzlibrary{calc}
\usetikzlibrary{fit}
\tikzstyle{block} = [rectangle,rounded corners, minimum width = 0cm, minimum height=1cm,text centered,inner sep=0.5mm] 
\tikzstyle{mytext} = [rectangle,rounded corners, minimum width = 0cm, minimum height=0cm,text centered] 

\newcommand{\y}{1.25}
\newcommand{\yy}{\y*2}
\newcommand{\x}{0.75}
\newcommand{\xx}{\x*2}
\newcommand{\thickness}{thick}

\newcommand{\scolor}{magenta}
\newcommand{\tcolor}{BlueGreen}
\newcommand{\eqcolor}{orange}
\def\inneralpha{10}
\def\tikzlinewidth{0.25}

\def\antennaDistance{0.25} 
\def\antennaYOffset{1.3}   
\def\antennaSpacing{0.5}  
\def\antennaSize{0.2}      
\def\labelOffset{0.25}    
\def\figurescaling{0.8}

\begin{document}

\title{RIS-aided Latent Space Alignment for Semantic Channel Equalization}

\author{Tomás Hüttebräucker$^{1}$, Mario Edoardo Pandolfo$^{2,3}$, Simone Fiorellino$^{2,3}$,\\ Emilio Calvanese Strinati$^1$, and Paolo Di Lorenzo$^{3,4}$ \medskip \\
$^1$ CEA Leti, University Grenoble Alpes, 38000, Grenoble, France.\\
$^2$ DIAG Department, Sapienza University of Rome, via Ariosto 25, Rome, Italy.\\
$^3$ Consorzio Nazionale Interuniversitario per le Telecomunicazioni (CNIT), Parma, Italy.\\
$^4$ DIET Department, Sapienza University of Rome, Via Eudossiana 18, Rome, Italy. \smallskip\\
E-mail: \{tomas.huttebraucker, emilio.calvanese-strinati\}@cea.fr, \\ \{marioedoardo.pandolfo, simone.fiorellino, paolo.dilorenzo\}@uniroma1.it.
\vspace{-.3cm}
}



\maketitle

\begin{abstract}
Semantic communication systems introduce a new paradigm in wireless communications, focusing on transmitting the intended meaning rather than ensuring strict bit-level accuracy. These systems often rely on \acp{DNN} to learn and encode meaning directly from data, enabling more efficient communication. However, in multi-user settings where interacting agents are trained independently—without shared context or joint optimization—divergent latent representations across AI-native devices can lead to \textit{semantic mismatches}, impeding mutual understanding even in the absence of traditional transmission errors. In this work, we address semantic mismatch in \ac{MIMO} channels by proposing a joint physical and semantic channel equalization framework that leverages the presence of \ac{RIS}. The semantic equalization is implemented as a sequence of transformations: (i) a pre-equalization stage at the transmitter; (ii) propagation through the RIS-aided channel; and (iii) a post-equalization stage at the receiver. We formulate the problem as a constrained \ac{MMSE} optimization and propose two solutions: (i) a linear semantic equalization chain, and (ii) a non-linear \ac{DNN}-based semantic equalizer. Both methods are designed to operate under semantic compression in the latent space and adhere to transmit power constraints. Through extensive evaluations, we show that the proposed joint equalization strategies consistently outperform conventional, disjoint approaches to physical and semantic channel equalization across a broad range of scenarios and wireless channel conditions.

\end{abstract}


\section{Introduction}

\IEEEPARstart{F}{or} the last seven decades, communication systems have been designed with the main objective of reliably transmitting symbols through noisy communication channels, typically disregarding the interpretation and impact of these symbols upon reception. 
Following this principle, communication networks have achieved significant advancements in bit transmission rate and reliability, fundamental metrics for data-centric applications such as video and audio streaming, where communication itself is the primary objective. However, many emerging \ac{6G} applications, including \ac{IoT}, smart cities, and autonomous vehicles, do not treat communication as an end but rather as a means to enable interaction and coordination among intelligent agents \cite{de2021survey}. This paradigm shift in the utilization of communication systems calls for a corresponding transformation in their design, moving the focus from optimizing accurate data transmission to maximizing the relevance and utility of the transmitted information for users' decision-making processes. In this context, \ac{SemComs} have emerged as a promising new paradigm for the design of communication networks \cite{strinati20216g,gunduz2022beyond,strinati2024goal}. Unlike conventional communication systems, which aim for perfect data recovery, the \ac{SemComs} paradigm focuses on preserving the underlying meaning of the transmitted data, relevant to a given task at the receiver. By discarding irrelevant information, \ac{SemComs} can potentially reduce communication overhead while maintaining, or even enhancing, downstream task performance. 


First introduced by Weaver in 1949 \cite{weaver1949recent} and formally addressed by Carnap and Bar-Hillel in 1952 \cite{carnap1952outline}, the concept of \ac{SemComs} traces back to the early development of digital communications. Despite its long-standing origins, \ac{SemComs} attracted limited attention from the scientific community in subsequent decades and was largely regarded as a theoretical or philosophical construct. In recent years, however, the widespread deployment of interconnected devices—threatening to overwhelm the capacity of communication networks—has renewed interest in the development of a theory of \ac{SemComs}~\cite{bao2011towards,guler2018semantic}, albeit with limited success. Practical realizations of \ac{SemComs} systems have also remained limited, primarily due to the inherent difficulties in formally defining, extracting, and interpreting semantic content from real-world data. Recent advancements in \ac{AI}, particularly in \ac{DL}, have addressed many of these challenges by enabling the development of \ac{SemComs} systems capable of learning a \textit{semantic protocol}—a structured method for extracting, transmitting, and reconstructing task-relevant meaning from data. This is typically achieved by leveraging \acp{DNN} to model semantic encoding and decoding functions, with \ac{ML} algorithms guiding their training for specific tasks. For instance, in semantic source coding \acp{DNN} are used to compress data based on an imperfect but semantically equivalent reconstruction. This approach has been shown to outperform traditional source coding methods both in compression and performance on the objective task \cite{talli2024effective,dubois2021lossy,binucci2024opportunistic}. However, semantic source coding does not account for the characteristics of the communication channel, which are fundamental to achieving efficient and reliable transmission. To address this limitation, \ac{JSCC} has emerged as an alternative that simultaneously considers both task performance and channel conditions. When implemented using \acp{DNN}, this approach is known as \ac{DJSCC} and has been widely explored in the \ac{SemComs} literature. First introduced in \cite{bourtsoulatze2019deep}, \ac{DJSCC} integrates source coding, channel coding, and modulation into a single end-to-end trainable \ac{DNN}-based model optimized for either reconstruction quality or task performance. This approach has demonstrated advantages over conventional communication systems across various applications, including image \cite{bourtsoulatze2019deep}, text \cite{xie2021deep,sana2022learning}, and audio \cite{weng2021semantic} transmission, as well as task-oriented scenarios such as image classification \cite{shao2021learning,beck2023semantic}, token communication \cite{devoto2025adaptive}, and sequential decision-making in \acp{MDP} \cite{tung2021effective}. Furthermore, its flexibility has facilitated its application to complex communication settings, including \ac{MIMO} \cite{wu2024deep}, \ac{OFDM} \cite{yang2022ofdm}, and feedback \cite{kurka2020deepjscc} channels.


While \ac{DNN}-based \ac{SemComs} systems offer notable advantages over traditional communication networks, they also introduce several challenges. Among these, one of the most significant is \textit{semantic noise}, which refers to all distortions that lead to misinterpretation of meaning, independently of the syntactic (physical) channel noise \cite{bao2011towards,luo2022semantic}. Semantic noise can arise from various sources, each with distinct characteristics and corresponding mitigation strategies. One such source is the inherent limitation of \ac{DL} models, which may produce inaccurate outputs due to restricted model capacity, limited training data, or poor generalization to unseen inputs. Closely related is adversarial noise \cite{hu2023robust}, which consists of small, semantically imperceptible perturbations deliberately designed to mislead the model—often leading to substantial performance degradation. While these sources of semantic noise have been studied and continue to receive attention in both the \ac{SemComs} and \ac{ML} literature, only recently a new source of semantic noise, \textit{semantic mismatch}, has been identified \cite{sana2023semantic}. This type of noise occurs when the transmitter and receiver interpret meaning differently—often due to discrepancies in their methods of extracting and interpreting semantic information. In \ac{DNN}-based \ac{SemComs}, such mismatch typically stems from inconsistencies in the learned latent representations between the transmitter and receiver, which typically occurs when their modules are not jointly trained. Semantic mismatch is often overlooked since most works assume that a joint training process across both ends of the communication link is possible when designing a \ac{SemComs} system. This assumption, however, poses a major limitation for real-world, multi-user deployments of \ac{SemComs}, where agents may use different neural architectures, datasets, or task objectives. In reality, such discrepancies are the norm—especially in multi-vendor environments where stakeholders are either unwilling or unable to share models, training data, or other proprietary resources. In these settings, joint end-to-end training or exchanging deep neural network models is often impractical due to privacy concerns and intellectual property restrictions. As a result, establishing a robust semantic alignment mechanism is not just beneficial—it is essential for ensuring consistency and interoperability across heterogeneous, real-world \ac{SemComs} systems.

A straightforward solution to semantic mismatch is joint fine-tuning, where all interacting agents briefly re-train their models to align communication protocols \cite{choi2024semantics}. However, this approach presents two major drawbacks. First, the energy and communication costs associated with re-training multiple \ac{DNN} models can outweigh the practical benefits of \ac{SemComs}. Second, agents may still be unable or unwilling to modify their models due to compatibility issues, operational constraints, or the aforementioned privacy concerns—making collaborative fine-tuning infeasible in many real-world scenarios. As an alternative to this, the concept of semantic channel equalization has been recently proposed \cite{sana2023semantic}. Inspired by traditional channel equalization techniques, this approach introduces the \textit{semantic channel equalizer}—a module, potentially distributed between the transmitter and receiver—that transforms semantic symbols to ensure consistent interpretation. 

Although still a relatively new concept, semantic channel equalization has already gained significant attention, with several studies exploring its design and practical implementation. In \cite{sana2023semantic}, the authors propose a semantic channel equalization module based on the concept of semantic space partition. Briefly, a partition of the semantic space (i.e., the space where semantic symbols are represented) is a grouping of its elements into non-empty and non-overlapping subsets, called atoms. Each atom in the partition is associated with a meaning or concept of interest. Based on the semantic space partition, they leverage a set of linear transformations tailored to align corresponding atoms transmitter and receiver. This approach has been extended to Markov Decision Process (MDP) scenarios \cite{huttebraucker2024pragmatic}, where self-supervised learning of the semantic partitioning has also been explored \cite{huttebraucker2024soft}. Alternatively, the concept of \ac{RR} \cite{moschella2022relative} has been utilized for semantic alignment assuming errorless communication. This approach establishes a common semantic space by encoding information relative to a shared subset of data, known as anchors. Notably, this shared space remains largely consistent across independently trained models and tasks. In \cite{fiorellino2024dynamic}, they leverage \ac{RR} to unify the semantic representation spaces of multiple encoders. Using a set of trained relative decoders, they perform dynamic optimization of the encoding model and network resources for efficient task completion. In \cite{huttebraucker2024relative}, they propose a \ac{RR}-based semantic channel equalization that requires no relative decoder and works with a small number of data samples. However, all the aforementioned works in semantic channel equalization either rely on a digital communications scheme or parallel analog \ac{SISO} channels for transmission, effectively separating semantic channel equalization from physical channel equalization.

\textbf{Contributions:} In this work, we present a novel methodological equalization framework for \ac{SemComs}, integrating both physical and semantic channel equalization in a \ac{MIMO} communication setting. Additionally, we investigate the potential of \ac{RIS} for semantic channel equalization. While recent studies have explored \ac{RIS} in the context of \ac{SemComs} \cite{jiang2024ris,hu2024drl,zhao2024joint,ma2024enhanced,shi2023reconfigurable}, its specific application to semantic channel equalization remains unexplored. Building on our preliminary conference work \cite{pandolfo2025latent}, and extending it to incorporate the presence of RIS in the semantic communication scenario, our equalization approach is structured into three stages: a pre-equalizer at the transmitter, a post-equalizer at the receiver, and a \ac{RIS}-based equalizer at the wireless the channel. We formulate the equalization problem as a constrained \ac{MMSE} problem, where constraints are imposed on the \ac{RIS} parameters and transmitted power. Moreover, by varying the communication budget, our formulation incorporates also a semantic compression of the latent space. We propose both linear and \ac{DNN}-based solutions. In the linear approach, the problem is decomposed into three subproblems, one for each set of variables, and solved through alternating optimization. The neural approach, on the other hand, is addressed using standard stochastic gradient descent algorithms. Moreover, we investigate the impact of various \ac{RIS} sizes, focusing on the trade-off between performance, compression, and complexity. Finally, we explore the potential of \ac{RIS} for adaptability to channel variations in \ac{SemComs}, where only the \ac{RIS} values are updated to enable successful semantic communications under channel variations.



In summary, the main contributions of our work are:
\begin{enumerate}
    \item We introduce a novel Multi-User \ac{SemComs} system model including semantic mismatch, \ac{MIMO} channel and \ac{RIS}. Following this model, we propose a holistic approach to perform joint compression as well as physical and semantic channel equalization. We pose the equalization problem as an MMSE optimization problem for which we obtain a Linear and a \ac{DNN}-based solution.

    \item We explore the potential of \ac{RIS} as a channel equalizer in the proposed scenario. Showing not only its capability to equalize channel changes but also to reduce semantic mismatch by introducing extra computation and optimization parameters.
    
    \item We compare our solutions extensively against baselines, showing their superior performance and exploring the tradeoff between compression, \ac{SNR}, and performance.

\end{enumerate}

\noindent\textbf{Outline.} The rest of the paper is organized as follows. In \cref{sec:system_model}, we introduce the multi-user semantic communications framework, highlighting the problem of semantic mismatch and formulating the semantic channel equalization problem. In \cref{sec:linear_solution} and \cref{sec:neural_solution}, we present the proposed linear and \ac{DNN}-based solutions. In \cref{sec:results}, we explain the simulation setup and results and finally, in \cref{sec:conclusions}, we draw the conclusions of our work.


\begin{figure*}[t]
    \centering
    \begin{adjustbox}{width=0.98\textwidth,center}
    \input{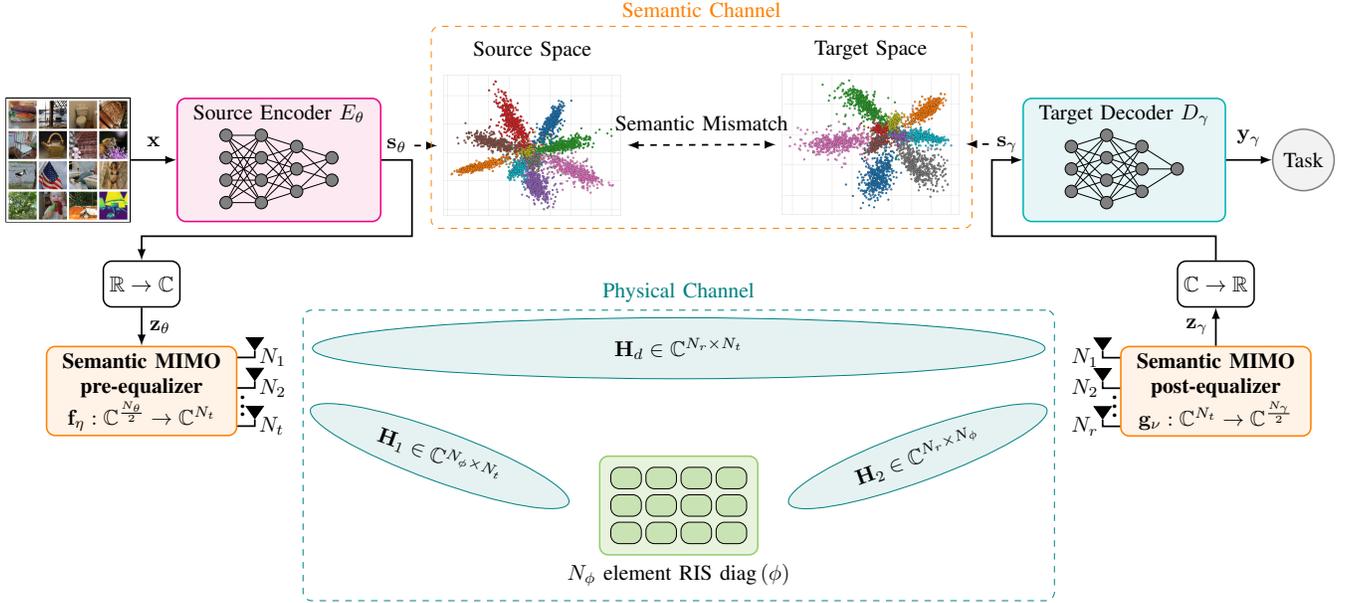}
    \end{adjustbox}
    \caption{Multi-User Semantic Communications with \ac{RIS}-empowered \ac{MIMO} communications.}
    \label{fig:semantic_equalization}
\end{figure*}

\section{System Model}\label{sec:system_model}

We consider the semantic communication framework depicted in \cref{fig:semantic_equalization}, where a transmitter and a receiver communicate by leveraging \ac{DNN}-based semantic encoding and decoding modules. In the single-user case, the transmitter and the receiver can jointly learn the encoding $\senc:\dspace\to\sspace$ and decoding $\sdec:\sspace\to\cY_\sw$ functions to enable effective communication. 
Here, $\sw$ denotes the shared model architecture and training process for both $\senc$ and $\sdec$. The set $\dspace$ represents the input data space, $\cY_\sw$ is the (agent-specific) action space, and $\sdim$ specifies the dimensionality of the semantic (latent) space to which semantic symbols are mapped.   
Then, given an observation $\data\in\dspace$, the transmitter extracts a semantic feature vector, $\slatentr=\senc\parenthesis{\data}\in\sspace$, which is transmitted to the receiver. Upon reception, the receiver utilizes $\sdec$ to infer an action, $\action_\sw$, from the set of possible actions $\cY_\sw$, based on the task at hand.

In multi-user semantic communications, the transmitter and receiver may employ independently learned communication strategies. Specifically, while the transmitter encodes the data $\data$ using the aforementioned $\senc$, the receiver might employ a different decoder $\tdec:\tspace\to\cY_\tw$ (trained in conjunction with a corresponding encoder $\tenc:\dspace\to\tspace$) for interpretation. This introduces a critical challenge: due to the independent training, naive transmission becomes prone to semantic interpretation errors resulting from a representation mismatch. That is, for the same data $\data$, the transmitted $\slatentr$ signal may not align with the adequate symbol for the receiver, represented by $\tlatentr=\tenc\parenthesis{\data}$. This mismatch can arise from many sources, such as differences in the training data distribution or task goals ($\cY_\tw\neq\cY_\sw$) and mismatches in the architectures ($\tdim\neq\sdim$). Even when the training data, the optimization objective, and the architecture are fixed for both training procedures, the inherent stochasticity of \ac{ML} algorithms will likely cause a discrepancy in the learned protocols \cite{moschella2022relative}.

In this work, we focus on \textit{equalizing} the \textit{semantic channel}, the underlying cause of semantic mismatch, jointly with the physical channel (hereafter referred to simply as the channel). For this, we consider a multi-user semantic communication scenario with a physical \ac{MIMO} channel with $N_t$ transmit and $N_r$ receive antennas, and a parallel diagonal \ac{RIS} channel with $N_\ris$ elements. The first step in designing the equalizer is to adapt the source semantic symbols into a complex representation. Assuming that $\sdim$ is even (otherwise, $\slatentr$ can be zero-padded), we define $\slatent \in \sspaceC$ as a complex vector whose real and imaginary parts are formed by splitting $\slatentr$ into two equal halves. We assume that, for each $\slatent$, the transmitter can use the channel $K$ times leveraging a pre-equalizer $\f_\eta:\sspaceC\to\bC^{KN_t}$ that maps the semantic representation $\slatent$ into $K N_t$ complex channel symbols. We model the channel as comprising a direct line-of-sight component, denoted by $\H_d \in \bC^{N_r \times N_t}$, and a reflective component induced by the \ac{RIS}, given by $\H_\ris = \H_2 \rism \H_1$, where $\rism$ is a diagonal matrix representing the RIS reflection coefficients. Here, $\ris=\{\ris_1,\ris_2,\ldots,\ris_{N_\ris}\}\in\bC^{N_\ris}$ is the array of phase shifts introduced by the \ac{RIS}, $\H_1$ denotes the channel from the transmitter to the \ac{RIS} and $\H_2$ the channel from the \ac{RIS} to the receiver. The \ac{RIS} enables successful communication in cases where the main channel $\H_d$ is attenuated. At the receiver side, the post-equalizer $\g_\nu:\bC^{KN_r}\to\tspaceC$ estimates the receiver's semantic symbol $\hat{\latent}_\tw$ , which can be converted back to a real representation $\hat{\latentr}_\tw$ by concatenating its real and complex parts. The modules $\f_\eta$ and $\g_\nu$ are modeled as functions with tunable parameters $\eta$ and $\nu$ respectively. They not only do physical and semantic channel equalization but also perform semantic compression of the transmitted latent vectors. We define the compression factor achieved by $\f_\eta$ and $\g_\nu$ as the ratio of the number of transmitted symbols and the total number of components of the transmitter symbol $\slatent$, i.e.:
\begin{equation}\label{eq:sem_compression_factor}
    \zeta=\frac{K N_t}{\frac{\sdim}{2}}.
\end{equation}
With the given model, the decoded signal at the receiver is:
\begin{equation}\label{eq:channel_model}
    \hat{\latent}_\tw = \g_\nu\parenthesis{\H_e^{\ris}\f_\eta\parenthesis{\slatent}+\noise},
\end{equation}
where $\noise\sim\mathcal{CN}\parenthesis{0,\Sigma_\noise}$ is the $KN_r$ dimensional Gaussian noise at the receiver and
\begin{equation}\label{eq:ris_channel}
\H_e^{\ris}=\I_K\otimes \parenthesis{\H_d+\H_2 \rism \H_1}\in\bC^{KN_r\times KN_t}    
\end{equation}
 is the $K$ usage effective channel matrix, with $\I_K$ denoting $K\times K$ identity matrix and $\otimes$ the Kronecker product. 

We formulate the channel-aware semantic alignment problem as an \ac{MMSE} optimization task, which aims to minimize the average euclidean distance between the latent spaces of the transmitter and the receiver, based on the model in \eqref{eq:channel_model}, and subject to constraints on the transmit power and the modulus of the \ac{RIS} phase shifts. Mathematically, this can be expressed as:
\begin{subequations} \label{eq:main_formulation}
    \begin{align}
        \min_{\eta,\nu,\ris}\quad &\bE_{\slatent,\tlatent,\noise}\left[\norm{\tlatent-\g_\nu\left(\H_e^{\ris}\f_\eta\left(\slatent\right)+\noise\right)}^2\right] \\
        \text{s.t.} \quad &\bE_{\slatent}\left[\norm{\f_\eta\left(\slatent\right)}^2\right] \leq P \label{eq:f_cond} \\
         &|\ris_i| = 1 \quad \forall i \in \{1, 2, \ldots, N_\ris\}, \label{eq:ris_cond}
    \end{align}
\end{subequations}
where $P$ is the available power budget at the transmitter. Condition \cref{eq:f_cond} assures that the average transmitted power does not exceed the budget, and condition \cref{eq:ris_cond} ensures that the \ac{RIS} parameters are passive and do not amplify the signal. Solving problem \cref{eq:main_formulation} requires the knowledge of the joint distribution of $\slatent$ and $\tlatent$, which is, in general, unknown. Thus, we propose replacing the expectation term with an empirical estimation using a data-driven approach. Assuming to have access to a limited dataset of $N$ samples
\begin{equation}\label{eq:dataset}
\dataset = \{\data^{(1)}, \data^{(2)}, \ldots, \data^{(N)}\} \subset \dspace^N,
\end{equation}
we define $\slatentri=\senc(\data^{(i)})$ and $\tlatentri = \tenc(\data^{(i)})$ as the real-valued latent vectors associated with the transmitter and receiver, respectively. We refer to each pair $\parenthesis{\slatentri, \tlatentri}$ as a \emph{semantic pilot}, which serves as a training anchor between the transmitter and receiver latent spaces to enable semantic alignment. In practice, semantic pilots may be obtained through a supervised calibration phase, where both the transmitter and receiver have access to a shared dataset of data samples. This dataset can be public, ensuring that no agent is required to disclose private data. Furthermore, since labels are not strictly necessary, the complexity of data collection is significantly reduced. The semantic pilots  $\parenthesis{\slatentri, \tlatentri}$ are then mapped to their corresponding complex-valued representations, denoted by $\parenthesis{\slatenti,\tlatenti}$, for transmission and processing.
Thus, we can reformulate \cref{eq:main_formulation} as the following data-driven approximation:
\begin{equation}\label{eq:data_driven_formulation}
\begin{aligned}%
        \min_{\eta,\nu,\ris}\quad & \sum_{i=1}^N\bE_{\noise}\left[\norm{\tlatenti-\g_\nu\left(\H_e^{\ris}\f_\eta\left(\slatenti\right)+\noise\right)}^2\right] \\
        \text{s.t.} \quad &\frac{1}{N}\sum_{i=1}^N\left[\norm{\f_\eta\left(\slatenti\right)}^2\right] \leq P \\
         &|\ris_i| = 1 \quad \forall i \in \{1, 2, \ldots, N_\ris\}.
\end{aligned}    
\end{equation}
In the sequel, we present solutions to problem \eqref{eq:data_driven_formulation}, based on a linear semantic equalizer approach in Section~\ref{sec:linear_solution}, and neural equalizers in Section~\ref{sec:neural_solution}.

\section{Linear Semantic Channel Equalization}\label{sec:linear_solution}
We first propose to address the channel-aware semantic alignment using a linear approach. We introduce the linear pre-equalizer $\F\in\bC^{KN_t\times\sdimC}$ and the linear post-equalizer $\G\in\bC^{\tdimC\times KN_r}$ to be jointly optimized with the RIS parameters $\ris$. Defining the data matrices $\sdata\in\bC^{\sdimC\times N}$ and $ \tdata\in\bC^{\tdimC\times N}$, where the $i$-th columns of $\sdata$ and $\tdata$ are the semantic pilots $\slatenti$ and $\tlatenti$ respectively, problem (\ref{eq:data_driven_formulation}) boils down to:
\begin{equation}
\begin{aligned}\label{eq:linear_formulation_data_driven}
    \min_{\F,\G,\ris}\quad &\frac{1}{N}\norm{\tdata-\G \H_e^{\ris}\F \sdata}^2_F+\operatorname{Tr}\parenthesis{\G^H\G\Sigma_\noise} \\
    \text{s.t.} \quad  &\operatorname{Tr}\parenthesis{\F^H\F\Sigma_{\slatent}}\leq P  \\
  &|\ris_i|=1 \quad \forall i\in\{1,2,\ldots,N_\ris\}.
\end{aligned}
\end{equation}
Here, $ \Sigma_{\slatent} $ denotes the covariance matrix of the latent variable $\slatent$, and we have used the fact that the noise $\noise$ is independent of the signal to move the noise term outside the expectation. Without loss of generality, we assume that the covariance matrix of the input signal, $\Sigma_{\sdata}$, is the identity matrix. This assumption can be satisfied through standard pre-whitening techniques, which apply a linear transformation to the data. Importantly, this transformation does not impact the expressiveness or capacity of the proposed equalizer.

Problem \cref{eq:linear_formulation_data_driven} is non-convex, making the optimal solution difficult to obtain. However, due to its three variable structure and its bi-convexity with respect to $\F$ and $\G$, it is well-suited for an alternating and iterative optimization approach. We therefore propose to exploit an \ac{ADMM}-based algorithmic soultion \cite{boyd2011distributed}, for which it is useful to recast \cref{eq:linear_formulation_data_driven} in an equivalent formulation:
\begin{equation}
\begin{aligned}
    \label{eq:linear_formulation_data_driven_admm}
    \min_{\F,\G,\ris} & \quad \frac{1}{N}\norm{\tdata-\G \H_e^{\ris}\F \sdata}^2_F+ \operatorname{Tr}\left(\G^H\G\right) + \operatorname{I}_{\mathcal{P}}\left(\R\right) \\ 
    \text{s.t.}& \quad  \F - \R = 0 \\
    &\quad |\ris_i|= 1 \quad \forall i \in \{1, 2, \dots, N_\ris\}.
\end{aligned}
\end{equation}
Here, $\operatorname{I}_{\mathcal{P}}$ is the indicator function for matrices belonging to set 
$\mathcal{P}=\{\R \,|\, \operatorname{Tr}\parenthesis{\R ^H\R}\leq P\}$, which is defined as:
\begin{equation}
    \operatorname{I}_{\mathcal{P}}\parenthesis{\R}=
    \begin{cases}
    +\infty &\text{if}\quad \operatorname{Tr}\parenthesis{\R ^H\R}>P\\
    0&\text{if}\quad \operatorname{Tr}\parenthesis{\R ^H\R}\leq P.
    \end{cases}
\end{equation}
The scaled ADMM solution to problem \cref{eq:linear_formulation_data_driven_admm} involves the following iterative alternating updates of the variables \cite{boyd2011distributed}:
\begin{align}
    &\begin{aligned}   
        \G^{\parenthesis{k+1}} &= \argmin_\G \frac{1}{N}\norm{\tdata- \G\H_e^{\ris^{\parenthesis{k}}}\F^{\parenthesis{k}}\sdata}^2   
        \\
        &+\operatorname{Tr}\parenthesis{\G^H\G\Sigma_\noise} 
    \end{aligned}\label{eq:G_obj}
    \\
    &\begin{aligned}
        \F^{\parenthesis{k+1}} &= \argmin_\F \frac{1}{N}\norm{\tdata-\G^{\parenthesis{k+1}}\H_e^{\ris^{\parenthesis{k}}}\F\sdata}^2 
        \\
        &+ \rho \norm{\F-\R^{\parenthesis{k}}+\S^{\parenthesis{k}}}^2 
    \end{aligned}\label{eq:F_obj}    
    \\
    &\R^{\parenthesis{k+1}} = \operatorname{Proj}_{\mathcal{P}}\parenthesis{\F^{\parenthesis{k+1}}+\S^{\parenthesis{k}}} \label{eq:R_obj}
    \\
    &\begin{aligned}
        \ris^{\parenthesis{k+1}} &= \argmin_\ris \norm{\tdata-\G^{\parenthesis{k+1}}\H_e^{\ris}\F^{\parenthesis{k+1}}\sdata}^2
        \\
        &\qquad\text{s.t.} \quad \abs{\ris_i}= 1 \quad \forall i \in \{1,2, \ldots,N_\ris\}
    \end{aligned}\label{eq:ris_obj}
    \\
    &\S^{\parenthesis{k+1}} = \S^{\parenthesis{k}}+\F^{\parenthesis{k+1}}-\R^{\parenthesis{k+1}}    
    \\    &\H_e^{\ris^{\parenthesis{k+1}}}=\I_K\otimes\parenthesis{\H_d + \H_2\text{diag}\parenthesis{\ris^{\parenthesis{k+1}}} \H_1}
\end{align}
where $k$ denotes the iteration step, $\operatorname{Proj}_{\mathcal{P}}(\cdot)$ denotes the orthogonal projection onto set $\mathcal{P}$, and $\rho>0$ is the regularization coefficient for the augmented Lagrangian \cite{boyd2004convex}. Each of the optimization steps in \cref{eq:G_obj}--\cref{eq:ris_obj} involves solving a smaller subproblem, whose solutions are detailed in the sequel.

\subsection{$\G$ step}
The step for $\G$ in \cref{eq:G_obj} can be solved in a closed form by setting the (conjugate) gradient $\nabla_{\G^H}$ of the objective to zero, thus yielding the following expression:
\begin{equation}
\begin{aligned}    
    \G^{\parenthesis{k+1}}&= \tdata \tilde{\ldata} ^H\parenthesis{\tilde{\ldata}\tilde{\ldata}^H+N \Sigma_\noise}^{-1},
\end{aligned}
\end{equation}
where $\tilde{\ldata}=\H_e^{\ris^{(k)}}\F^{\parenthesis{k}}\sdata$ is the signal received by the decoder in the absence of channel noise.

\subsection{$\F$ step}
Similarly, setting the conjugate gradient $\nabla_{\F^H}$ of \cref{eq:F_obj} to zero yields the expression of the optimal $\F$:
\begin{equation}\label{eq:F_step}
    \mathbf{A}\F^{(k+1)}+\F^{(k+1)}\mathbf{B}=\mathbf{C}
\end{equation}
where
\begin{align}
    & \mathbf{A} = \parenthesis{\G^{(k+1)}\H_e^{\ris^{(k)}}}^H\parenthesis{\G^{(k+1)}\H_e^{\ris^{(k)}}}
    \\
    & \mathbf{B} = N\rho\parenthesis{\sdata \sdata^H}^{-1} 
    \\
    & \mathbf{C} = N\rho\parenthesis{\R^{(k)}-\S^{(k)}}+\parenthesis{\G^{(k+1)}\H_e^{\ris}}^H\tdata \sdata^H.
\end{align}
Equation \cref{eq:F_step} is the well-known Sylvester Equation which can be efficiently solved for $\F^{(k+1)}$ using the Bartels-Stewart algorithm \cite{bartels1972algorithm}.

\subsection{$\R$ step}
The $\R$ step in (\ref{eq:R_obj}) requires a projection onto the set $\mathcal{P}$ of matrices with the required power constraint. 
Letting $\tilde{\R}=\F^{(k+1)}+\S^{(k)}$, we have:
\begin{equation}\label{eq:V_step}
 \begin{aligned}
    \R^{(k+1)} &= \operatorname{Proj}_{\mathcal{P}}\parenthesis{\tilde{\R}}
    \\
    &= \argmin_\R \quad \norm{\tilde{\R}-\R}_F^2\\
    &\ \qquad \text{s.t.} \quad \operatorname{Tr}\parenthesis{\R\R^H}<P.
\end{aligned}
\end{equation}
Problem \cref{eq:V_step} is a convex problem and can thus be solved optimally by imposing the \ac{KKT} conditions:
\begin{align}
    \nabla_{\R^H}\parenthesis{\norm{\tilde{\R}-\R}_F^2 + \lambda  \parenthesis{\operatorname{Tr}\parenthesis{\R\R^H}-P}} &= \mathbf{0} \label{eq:V_gradient}
    \\
    \operatorname{Tr}\parenthesis{\R\R^H}-P &\leq0
    \\
    \lambda &\geq0
    \\
    \lambda\parenthesis{\operatorname{Tr}\parenthesis{\R\R^H}-P}&=0.
\end{align}
From \cref{eq:V_gradient} we obtain 
\begin{equation}
    \R = \frac{1}{1+\lambda}\tilde{\R}.
\end{equation}
Knowing that $\operatorname{Tr}\parenthesis{\R\R^H}=\displaystyle\frac{1}{\parenthesis{1+\lambda}^2}\operatorname{Tr}\parenthesis{\tilde{\R}\tilde{\R^H}}$, the optimal $\lambda$ is given by
\begin{equation}
    \lambda^* = 
    \begin{cases}
    0,\qquad &\text{if } \operatorname{Tr}\parenthesis{\tilde{\R}\tilde{\R}^H}\leq P\\
    \sqrt{\operatorname{Tr}\parenthesis{\displaystyle\frac{\tilde{\R}\tilde{\R}^H}{P}}}-1 &\text{else.}
    \end{cases}
\end{equation}
Thus, the final update for $\R$ reads as:
\begin{equation}
    \R^{(k+1)} = \frac{1}{1+\lambda^*}\parenthesis{\F^{(k+1)}+\S^{(k)}}.
\end{equation}  


\subsection{$\ris$ step}
Updating the \ac{RIS} parameters involves solving the constrained optimization problem in \cref{eq:ris_obj}, which can be equivalently reformulated as the following constrained \ac{MSE} minimization problem:
\begin{equation}\label{eq:ris_problem_MSE}
\begin{aligned}
    \ris^{(k+1)} =  \argmin_\ris& \quad \norm{\v-\tilde{\D}\parenthesis{\mathbf{1}_K\otimes\ris}}^2_F
    \\
     \text{s.t.}&\quad \abs{\ris_i}= 1 \quad \forall i \in \{1,2, \ldots,N_\ris\},
\end{aligned}
\end{equation}
where $\mathbf{1}_K \in \bC^{K}$ denotes the $K$-dimensional vector of all ones, and the vector $\v \in \bC^{\tdimC N}$ and the matrix $\tilde{\D} \in \bC^{\tdimC N \times K N\ris}$ are defined as:
\begin{align}
    &\v= \text{vec}\parenthesis{\tdata-\G^{(k+1)}\parenthesis{\I_K\otimes\H_d}\F^{(k+1)}\sdata} \label{eq:vector_v}
    \\
    &\text{col}_j\parenthesis{\tilde{\D}} = \text{vec} \Big(
            \text{col}_j\parenthesis{\mathbf{P}} \text{row}_j\parenthesis{\mathbf{Q}}\Big),
        \label{eq:matrix_D_tilde}
\end{align}
with $\mathbf{P}=\G^{(k+1)}\parenthesis{\I_K\otimes\H_2}$ and $\mathbf{Q}=\parenthesis{\I_K\otimes\H_1}\F^{(k+1)}\sdata$. The operations $\text{col}_j$ and $\text{row}_j$ denote the selection of the $j$-th column and row, respectively. This way, $\text{col}_j\parenthesis{\mathbf{P}}\in\bC^{\tdimC\times 1}$, $\text{row}_j\parenthesis{\mathbf{Q}}\in\bC^{1\times N}$ and the product $\text{col}_j\parenthesis{\mathbf{P}} \text{row}_j\parenthesis{\mathbf{Q}}\in\bC^{\tdimC\times N}$. Finally, $\text{vec}\parenthesis{\cdot}$ denotes the vectorization operation which flattens a matrix into a vector by stacking its columns. Problem \cref{eq:ris_problem_MSE} can also be reformulated to avoid the Kronecker product, introducing the matrix $\D\in\bC^{\tdimC N\times N_\ris}$, whose elements are defined as:
\begin{equation}\label{eq:matrix_D}
    \squareb{\D}_{i,j} = \sum_{l=0}^{K-1} \squareb{\tilde{\D}}_{i,j+N_\ris l},
\end{equation}
where $[\cdot]_{i,j}$ denotes the element in the $i$-th row and $j$-th column of a matrix. Each column of $\D$ is the sum of the columns of $\tilde{\D}$ which correspond to the same $\ris$ element. Thus, problem \cref{eq:ris_problem_MSE} can be equivalently recast as: 
\begin{subequations}\label{eq:ris_problem_MSE_simple}
\begin{align}
    \ris^{(k+1)} =  \argmin_\ris& \quad \norm{\v-\D\ris}^2_F
    \\
     \text{s.t.}&\quad \abs{\ris_i}= 1 \quad \forall i \in \{1,2, \ldots,N_\ris\} \label{cond:ris_non_conv}.
\end{align}
\end{subequations}
Unfortunately, problem \cref{eq:ris_problem_MSE_simple} is still non-convex due to the constraint \cref{cond:ris_non_conv}, which makes the optimal solution difficult to obtain. Moreover, even if $\ris^{(k+1)}$ were optimal for \cref{eq:ris_problem_MSE_simple}, it would not necessarily be optimal for the overall problem defined in \cref{eq:linear_formulation_data_driven}, since the variables $\G$ and $\F$ are updated in subsequent steps. 
Therefore, we adopt an approximate solution using the iterative \ac{PGD} method. Rather than fully solving the subproblem at each iteration, we perform only one \ac{PGD} step to allow incremental adaptation of $\ris$. This approach is justified by the alternating nature of the main algorithm, where the subproblem for $\ris$ is redefined each iteration following updates to $\G^{(k+1)}$ and $\F^{(k+1)}$. Our goal at each step is not to find a locally optimal $\ris$, but to improve upon its previous value, ensuring steady progress with lower computational cost. Finally, defining the component-wise projection function
\begin{equation}
    \Pi(\ris_i) = 
    \frac{\ris_i}{\abs{\ris_i}}
\end{equation}
and the vector projection function
\begin{equation}\label{eq:projection}
    \mathbf{\Pi}(\ris)=\curlyb{\Pi(\ris_1),\Pi(\ris_2),\ldots,\Pi(\ris_{N_\ris})},
\end{equation}
the $\ris$ step reads as
\begin{equation}\label{eq:ris_step}
    \ris^{\parenthesis{k+1}} = \mathbf{\Pi} \parenthesis{\ris^{\parenthesis{k}} - \alpha \D^H\parenthesis{\D\ris^{\parenthesis{k}}-\v}},
\end{equation}
where $\D$ and $\v$ are defined in \cref{eq:matrix_D} and \cref{eq:vector_v} respectively.

\section{Neural Semantic Channel Equalization}\label{sec:neural_solution}
While the proposed linear solution benefits from low complexity, it has a limited capability to equalize non-linear mismatches between the source and target languages. Therefore, we propose to perform equalization based on \acp{DNN}. Specifically, we let $\f_\eta$ and $\g_\nu$ be Complex-\acp{DNN} for which the optimization in \cref{eq:main_formulation} is done over the network weights $\eta$ and $\nu$ respectively. Leveraging the dataset $\dataset$ defined in \cref{eq:dataset} an the corresponding semantic pilots $\parenthesis{\slatenti,\tlatenti}$ the \ac{DNN} loss function is 
\begin{equation}\label{eq:neural_formulation_data_driven}
    \mathcal{L}\parenthesis{\eta,\nu,\ris}= \frac{1}{N} \sum_{i=1}^{N}\squareb{\norm{\tlatenti-\g_\nu\parenthesis{\H_e^{\ris}\f_\eta\parenthesis{\slatenti}+\noise^{(i)}}}^2}.
\end{equation}
where, $\noise^{(i)}\sim\mathcal{CN}\parenthesis{0,\Sigma_\noise}$. Following traditional \ac{ML} approaches, the models are iteratively optimized using stochastic gradient-based methods where each $\noise^{(i)}$ is re-sampled in every iteration. The transmit power \cref{eq:f_cond}  and passive \ac{RIS} \cref{eq:ris_cond} constraints are directly included in the forward pass of the model. Specifically, the output of $\f_\eta$ is normalized and the effective channel is defined as
\begin{equation}
    \H_e^{\ris}=\I_K\otimes \parenthesis{\H_d+\H_2 \text{diag}\parenthesis{{\mathbf{\Pi}\parenthesis{\ris}}}\H_1},
\end{equation}
where $\mathbf{\Pi}$ is the projection defined in \cref{eq:projection}. These operations are captured by the gradients $\nabla_\eta \mathcal{L}$ and $\nabla_\ris \mathcal{L}$ and therefore, the optimization by stochastic gradient-based \ac{DL} algorithms is constraint-aware.


To improve the efficiency of the neural equalizer, we induce sparsity in the model's parameters \(\nu\) and \(\eta\) through hard thresholding. Specifically, after each training iteration, we apply a thresholding operator that sets to zero all weights whose magnitude is below a fixed value. This process explicitly enforces sparsity by removing low-magnitude parameters that are assumed to contribute minimally to the learned transformation. Given a sparsity factor \(\beta\) and learning rate $\tau$, we fix the pruning threshold at \(\beta \tau\). Once a weight is pruned, it remains zeroed for all subsequent iterations. The full training procedure, including thresholding and frozen updates, is summarized in \cref{alg:hard_thresholding_eta_nu}. After each training step, the set of pruned parameters is updated to include any newly sub-threshold weights, and all parameters in this set, including previously pruned ones, are explicitly zeroed out.

\begin{algorithm}[ht]
\caption{\ac{DNN} training algorithm with Hard Thresholding for sparsity}
\label{alg:hard_thresholding_eta_nu}
\begin{algorithmic}[1]
\Require Initial weights \(\eta_0, \nu_0, \ris_0\), learning rate \(\tau\), sparsity value \(\beta\)
\State Initialize pruning set $Z_0=\{\}$
\For{$t = 1$ to $T$}
    \State Compute loss: \(\mathcal{L}(\eta_{t-1}, \nu_{t-1}, \ris_{t-1})\)
    \State Update parameters: $\nu_t, \eta_t, \ris_t \gets \mathrm{Weight Update}(\nabla \mathcal{L})$
    \State Update pruning set: $Z_t=Z_{t-1}\cup\curlyb{w_i|\abs{w_i}<\beta\tau, w_i\in \{\eta_t,\nu_t\}}$
    \State Apply pruning: $w_i\gets0,\forall w_i\in Z_t$
\EndFor
\State \Return Final weights \(\nu_T, \eta_T, \ris_T\)
\end{algorithmic}
\end{algorithm}

\section{Simulation Setup and Results}\label{sec:results}
To evaluate our results, we choose the Multi-User Semantic Communications scenario depicted in \cref{fig:semantic_equalization}, where the transmitter uses $\senc$ to encode images and the receiver uses $\tdec$ to classify them.

\textbf{Wireless scenario:} Communication is established through a \ac{RIS}-aided square ($N_t=N_r)$ \ac{MIMO} channel. We adopt a Rayleigh fading model for the direct channel $\H_d$ and the \ac{RIS} parallel channels $\H_1$ and $\H_2$. Specifically, for channel $\H_\ell$ ($\ell\in\curlyb{d,1,2}$), each of its components is modeled as a complex Gaussian random variable with distribution $\mathcal{CN}\parenthesis{0,\alpha_\ell^{-1}}$ where $\alpha_\ell$ is the attenuation of channel $\H_\ell$ related to path-loss, blocking and other physical effects. We fix $\alpha_d=10^6$, $\alpha_1=\alpha_2=10$ unless stated otherwise. We set the noise covariance matrix to $\Sigma_\noise=\sigma_\noise^2\I_{KN_r}$, where $\I_{KN_r}$ is the $KN_r$ dimensional identity matrix and $\sigma_\noise^2$ is the noise power. The \ac{SNR} per sample is then defined as 
\begin{equation}
    \text{SNR}_\text{true} = \frac{\bE\squareb{ \norm{\H_e^\ris \f\parenthesis{\slatent} }^2}}{\bE\squareb{\norm{\noise}^2}}=\frac{N_r\alpha_e^{-1}P}{KN_r\sigma_\noise},
\end{equation}
where $\alpha_e$ is the average squared absolute value of each element of $\H_e^\ris$. The value of $\alpha_e$ will depend on the values of $\ris$ obtained by the proposed solution and can not be fixed. Therefore, to perform a fair comparison we fix the transmission \ac{SNR} instead, which we define as
\begin{equation}\label{eq:sim_snr}
    \text{SNR} = \frac{P}{K\sigma_\noise}.
\end{equation}
The \ac{SNR} decreases with $K$, which makes the comparison among different compression factor $\zeta$ values \cref{eq:sem_compression_factor} difficult. To isolate the compression effect, $P$ can be set proportional to $K$ or, equivalently, $\sigma_\noise$ inversely proportional to $K$. We choose for simulation convenience the latter approach and set $P=1$ and $\sigma_\noise=\frac{1}{K\text{SNR}}$ for a chosen value of \ac{SNR}.


\textbf{Task and architectures:} For the encoders, we selected two standard \ac{DNN} models from the \textit{timm} library \cite{rw2019timm}: \textit{vit\_base\_patch16\_224} for $\senc$ ($N_\sw=768)$ and \textit{vit\_small\_patch16\_224} for $\tenc$ ($N_\tw=384)$. Both models have been pre-trained on the ImageNet dataset \cite{deng2009imagenet} in a noiseless communication scenario. We evaluate the performance on the image classification dataset CIFAR10 \cite{krizhevsky2009learning} composed of $6\times 10^4$ $32\times32$ color images evenly distributed among 10 classes. We use $5\times 10^4$ samples to train the decoder $\tdec$ using the fixed encoder $\tenc$, and, of the remaining $10^4$ samples, we use $N$ train samples to optimize the proposed equalization models and $10^4-N$ for testing. For the neural equalizer, we select $N_\text{val}$ samples out of the $N$ training samples for validation. Unless otherwise specified, we set $N=5\times 10^3$ and $N_\text{val}=10^3$.  Both the architectures of $\f_\eta$ and $\g_\nu$ are formed with a single hidden-layer of $\tdimC=192$ complex neurons with a GELU activation \cite{hendrycks2016gaussian} on the complex magnitude. Unless specified otherwise, the sparsity factor $\beta$ (cf. Algorithm 1) is set to 0.

\textbf{Baselines:} Following \cite{pandolfo2025latent}, we compare our proposed methods with a disjoint alignment and transmission approach. The semantic alignment is done at the receiver leveraging a linear transformation $\T\in\bR^{\tdim\times\sdim}$ optimized to minimize the \ac{MSE} without communication. Using the dataset $\dataset$ defined in \cref{eq:dataset}, we create the data matrices $\sdatareal\in\bR^{\sdim\times N}$ an $\tdatareal\in\bR^{\tdim\times N}$, whose $i$-th columns are the semantic pilots $\slatentri$ and $\tlatentri$ (source and target real latent vectors associated to $\data^{(i)}$) respectively and we define $\T$ as
\begin{equation}
    \T = \argmin_{\T'\in\bR^{\tdim\times\sdim}} \norm{\T'\sdatareal-\tdatareal}^2_F.
\end{equation}
Each source symbol $\slatentr$ is first converted to a complex representation $\slatent$ and sent through the channel relying on analog transmission. The recovered symbol $\hat{\latent}_\sw$ is converted to a real representation $\hat{\latentr}_\sw$ and used to perform alignment $\hat{\latentr}_\tw=\T\hat{\latentr}_\sw$. For a fair comparison, for each sample we allow only the transmission of $KN_t$ channel symbols, or equivalently $2KN_t$ real components, same as our proposed methods. To select the components for transmission, we consider different strategies. 
The \textbf{Baseline First-$\kappa$} strategy consists of selecting the first $2KN_t$ components of $\slatentr$ to build $\slatent$, setting the remaining components to zero in reconstructions. The second strategy, \textbf{Baseline Top-$\kappa$}, consists of transmitting the components of $\slatentr$ with the largest absolute value. In order to recover the aligned vector, the receiver needs to know the index of the components transmitted with Baseline Top-$\kappa$, for this we assume that they are sent in parallel, meaning that $\frac{KN_t}{2}$ channel symbols are used to send $\slatent$ (equivalent to transmitting the top $KN_t$ components of $\slatentr$) and the remaining $\frac{KN_t}{2}$ are used to transmit the indices. For simulation simplicity, we assume that the indices are recovered perfectly. The final strategy is called \textbf{Baseline Eigen-$\kappa$} and consists of leveraging the \ac{SVD} of $\T=\U_\T\Sigma_\T \V_\T^T$ to perform a low-rank regression. For this we truncate the \ac{SVD} matrices to represent the $2KN_t$ largest singular values, obtaining  $\U_{\T_\kappa}$, $\Sigma_{\T_\kappa}$ and $\V_{\T_\kappa}^T$ respectively. We then define $\mathbf{e}_\kappa = \Sigma{\T_\kappa}^{1/2} \V_{\T_\kappa}^\top \slatentr$ as the compressed version of $\slatentr$, which is converted to its complex representation and transmitted through the channel. The recovered symbol is given by $\hat{\latentr}_\tw = \U{\T_\kappa} \Sigma_{\T_\kappa}^{1/2} \hat{\mathbf{e}}_\kappa$, where $\hat{\mathbf{e}}_\kappa$ denotes the estimate of $\mathbf{e}_\kappa$ after channel transmission, traditional equalization, and conversion back to the real domain. Contrary to First-$\kappa$ and Top-$\kappa$, the Eigen-$\kappa$ method performs joint compression and semantic equalization, this way, no further alignment step is necessary after the decompression.

For all baseline strategies, we perform the physical channel equalization jointly with the optimization of the \ac{RIS} parameters $\ris$. Starting from a random $\ris^{(0)}$ we do
\begin{align}
    &\H_e^{\ris^{\parenthesis{k+1}}}=\H_d + \H_2\text{diag}\parenthesis{\ris^{\parenthesis{k}}} \H_1 = \mathbf{U}\Sigma\mathbf{V}^H \\
    &\F^{(k+1)} = \frac{1}{\sqrt{N_t}}\mathbf{V} \\
    &\G^{\parenthesis{k+1}}=\sqrt{N_t}\parenthesis{\Sigma^H\Sigma+N_r\sigma_\noise^2 \I_{N_r}}^{-1}\parenthesis{\mathbf{U}\Sigma}^H \\
    &\begin{aligned}
        \ris^{\parenthesis{k+1}} &= \argmin_\ris \norm{\I_{N_r}-\G^{\parenthesis{k+1}}\H_e^{\ris}\F^{\parenthesis{k+1}}\I_{N_t}}^2
        \\
        &\qquad\text{s.t.} \quad \abs{\ris_i}= 1 \quad \forall i \in \{1,2, \ldots,N_\ris\}.
    \end{aligned}\label{eq:ris_baseline}
\end{align}
That is, at each iteration we first obtain the \ac{SVD} equalization of the \ac{RIS}-effective channel $\H_e^\ris$ and then we update the \ac{RIS} parameter to further improve the equalization. For problem \cref{eq:ris_baseline} we follow the same \ac{PGD} approach as in \cref{eq:ris_step} by appropriately redefining the vector $\v$ and the matrix $\D$. For all simulations, we perform a total of 30 iterations and set the final $\G$ and $\F$ as the \ac{SVD} \ac{MMSE} equalization of the channel with the final value of $\ris$.



\textbf{Simulations:} All results are averaged over 5 different random seeds used to generate the channels $\H_d$, $\H_1$, and $\H_2$, the sample shuffling to obtain $\sdata$ and $\tdata$, the decoders $\tdec$ and $\sdec$, and the initializations of $\F$, $\ris$ and weights $\nu$ and $\eta$ for each method. The reported curves represent the average values across these 5 runs. For each metric, we also indicate the range by displaying the highest and lowest values observed across the seeds. 


\subsection{Performance with varying compression factor}

\begin{figure*}
    \centering
    \begin{subfigure}[b]{\linewidth}
        \centering
        \includegraphics[width=\figurescaling\linewidth]{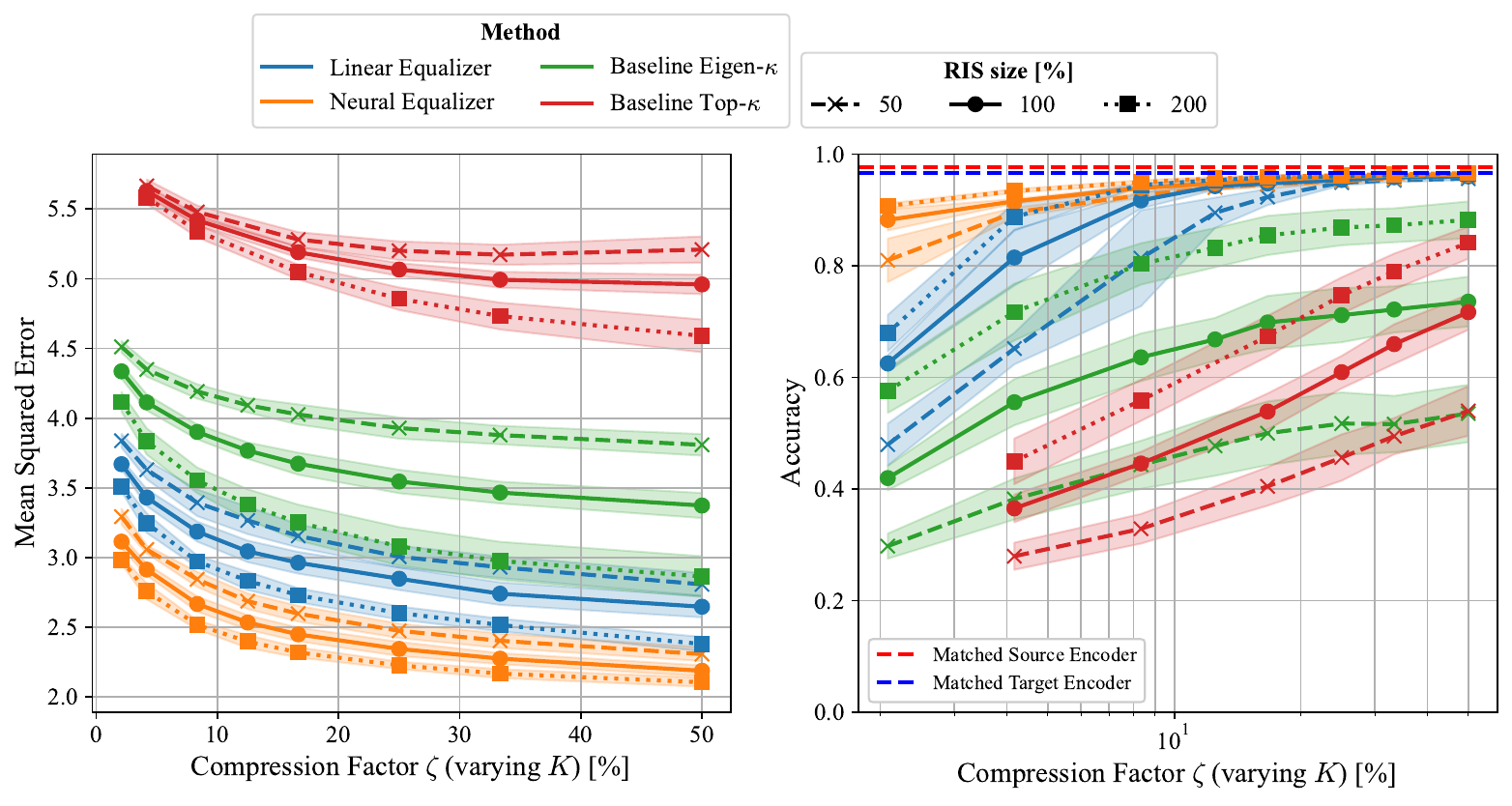}
        \caption{Varying $K$, with $N_t=8$. \ac{RIS} size is measured as $\frac{N_\ris}{N_t}$.}
        \label{fig:baseline_comparisson_K}
    \end{subfigure}
    \hfill
    \begin{subfigure}[b]{\linewidth}
        \centering
        \includegraphics[width=\figurescaling\linewidth]{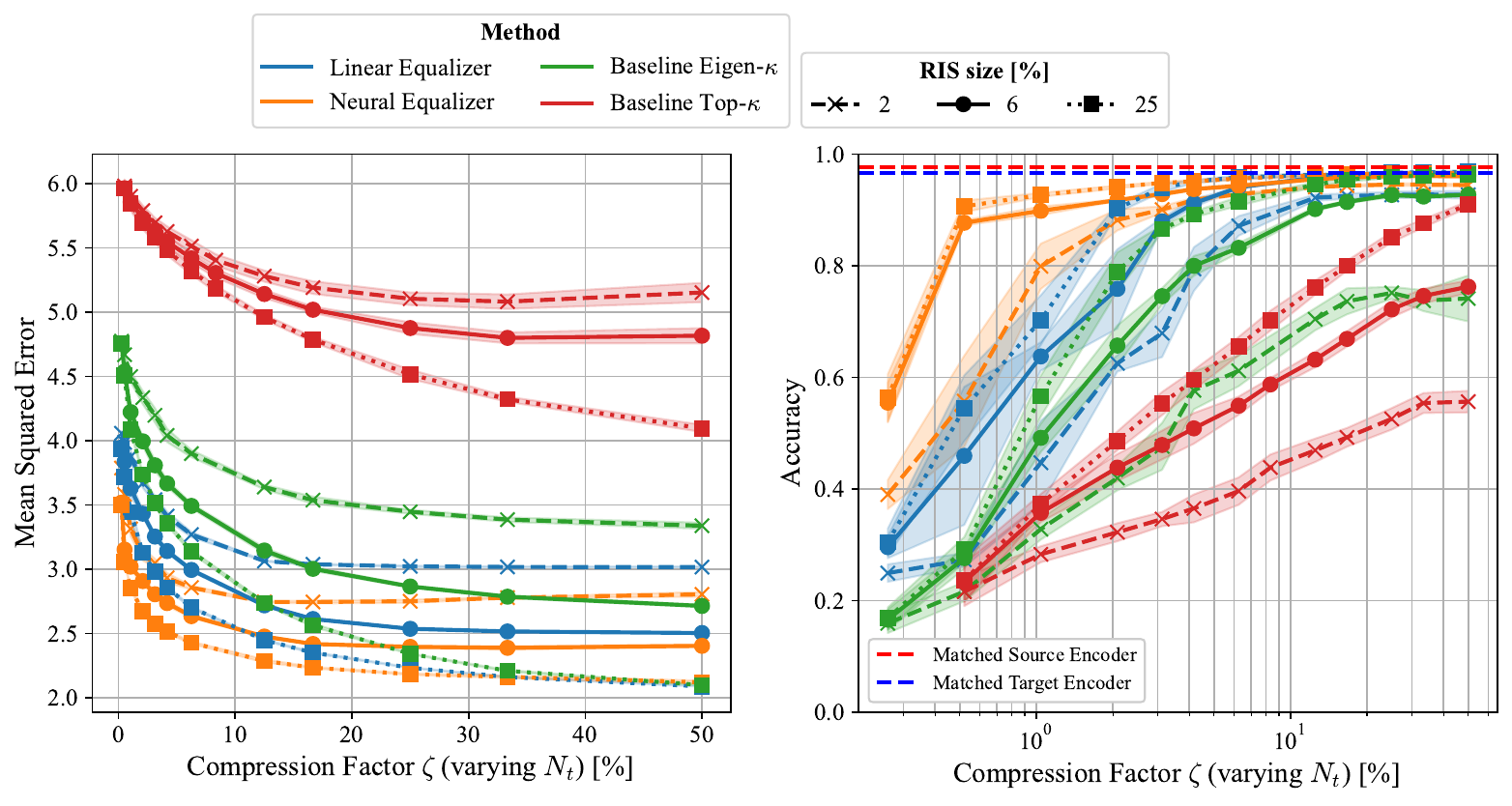}
        \caption{ Varying $N_t$, with $K=1$. \ac{RIS} size is measured as $\frac{2N_\ris}{\sdim}$}
        \label{fig:baseline_comparisson_Nt}
    \end{subfigure}
    \caption {Performance of the proposed Linear and Neural equalization methods compared with a separate alignment and transmission approach. Shared parameters: $\text{SNR}=10$ dB, $\alpha_d=10^{6}$, and $\alpha_1=\alpha_2=10$.}
    \label{fig:baseline_comparisson_combined}
\end{figure*}

We simulate the semantic communication scenario with different values of semantic compression factor $\zeta$, defined in \cref{eq:sem_compression_factor}, for $\text{SNR}=10$ dB. In \cref{fig:baseline_comparisson_combined}, we show the performance of the proposed linear and neural equalizers compared with the Eigen-$\kappa$  and Top-$\kappa$ baselines as a function of $\zeta$ for different \ac{RIS} sizes. To sweep $\zeta$, we employ two strategies. The first strategy, corresponding to \cref{fig:baseline_comparisson_K}, involves fixing $N_t=N_r=8$ and varying the number of channel uses $K$. The second strategy, corresponding to \cref{fig:baseline_comparisson_Nt}, consists of fixing $K=1$ and varying the size of the squared \ac{MIMO} channel $N_t$. In the varying $K$ strategy, we express the \ac{RIS} size as a fraction of the channel size, $\frac{N_\ris}{N_t}$. In contrast, in the varying $N_t$ strategy, we quantify the \ac{RIS} size as a proportion of the total latent space dimension, $\frac{2N_\ris}{\sdim}$. This allows for a direct comparison between the \ac{RIS} size and the effective compression factor, taking into account that the direct channel $\H_d$ is blocked and communication occurs through the \ac{RIS} channel. We show both the \ac{MSE} obtained by the equalization and the accuracy of the equalized models. Furthermore, we show as a benchmark the accuracy of the perfectly matched models with no compression and a noiseless communication channel. The \textbf{Matched Source Encoder} corresponds to the performance of $\senc$ paired with $\sdec$, and the \textbf{Matched Target Encoder} the performance of $\tenc$ paired with $\tdec$. 

\subsubsection{Comparison with Baselines}
As we can see from \cref{fig:baseline_comparisson_combined}, our proposed Linear and Neural equalization methods consistently outperform the baselines across all compression factor values and \ac{RIS} sizes. Since our methods perform joint compression and channel equalization—both semantic and physical—they are more effective at aligning the source and target languages while also accounting for the compression required for transmission. In contrast, the baselines perform compression in a communication-agnostic manner. In particular, the Top-$\kappa$ baseline applies compression without considering alignment or communication, while the Eigen-$\kappa$ method performs joint alignment and compression but remains communication-agnostic. This distinction explains why the Eigen-$\kappa$ method outperforms the Top-$\kappa$ approach in terms of \ac{MSE} across all $\zeta$ values and \ac{RIS} sizes. While this trend generally holds for the Accuracy as well, results show that the relationship between \ac{MSE} and Accuracy is not strictly direct. For example, in \cref{fig:baseline_comparisson_K}, for any $\zeta$ value, the Eigen-$\kappa$ method with the 50\% \ac{RIS} size consistently outperforms the Top-$\kappa$ method with 200\% \ac{RIS} size in terms of \ac{MSE}. However, this does not hold for Accuracy: the Top-$\kappa$ method with the highest \ac{RIS} size achieves higher Accuracy than the Eigen-$\kappa$ method with the smallest \ac{RIS} size, with a margin exceeding 20\% at the highest $\zeta$ value.

These results highlight that although \ac{MSE} has served as an effective optimization metric for both our proposed methods and the baselines, it does not always correlate well with downstream task Accuracy. In particular, the high de-correlation between these measures for the Top-$\kappa$ method suggests that it is the error of the largest components that is more relevant for accuracy, rather than the \ac{MSE}. This phenomenon, also observed in prior work \cite{huttebraucker2024relative}, motivates the exploration of alternative optimization metrics, such as a weighted version of the \ac{MSE}, that better align with task-specific performance. However, such investigations fall outside the scope of this paper and are left for future work.


\subsubsection{Effect of the compression factor $\zeta$}
Varying the compression factor $\zeta = \frac{2KN_t}{N_\sw}$ with respect to $K$ has two main effects on the proposed methods. First, a higher compression factor implies a larger number of parameters to optimize. For linear functions $\F$ and $\G$, the number of parameters is $N_\sw K N_t$ and $K N_r N_\tw$, respectively. In the case of the neural equalizer, the parameter count for $\f_\eta$ is $N_\tw (N_\sw + 1) + N_\tw (K N_t + 1)$, while for $\g_\nu$ it is $N_\tw (N_\tw + 1) + K N_r (N_\tw + 1)$. As such, increasing $\zeta$ enhances the expressive capacity of the models, thereby improving their ability to align the communication protocols. Secondly, $\zeta$ defines a bottleneck in the information flow between the encoder and decoder. Due to power constraints and channel noise, the number of transmitted symbols per image imposes a fundamental ceiling on the achievable information transfer. Therefore, the compression factor inherently limits the capacity, regardless of model complexity. As expected, as we can see from \cref{fig:baseline_comparisson_combined}, increasing $\zeta$ (which corresponds to higher model complexity and more transmitted symbols) leads to improved performance in the proposed methods.

\subsubsection{Effect of the \ac{RIS} size}
The \ac{RIS} size also affects the performance of the methods in multiple ways. Similarly to the semantic compression factor, the \ac{RIS} size defines a bottleneck in the communication. As the direct channel $\H_d\in\bC^{N_r\times N_t}$ is blocked, the information travels through the \ac{RIS}, effectively projected by $\H_1\in\bC^{N_\ris\times N_t}$ in a complex vector of $KN_\ris$ dimensions. On the other hand, the \ac{RIS} size will affect the overall norm of the effective channel $\H_e^\ris=\I_K\otimes\parenthesis{\H_d+\H_2 \text{diag}\parenthesis{\ris}\H_1}$. While each element in $\ris$ is fixed to an absolute value of 1, including more elements enables the transmission of a more powerful signal, which makes it more robust to noise. This is, indeed, the main idea behind the implementation of \ac{RIS}. Finally, even if less relevant compared to the pre and post equalizer, the \ac{RIS} also introduces more optimization parameters which can yield a performance gain given by increased complexity. From both \cref{fig:baseline_comparisson_K} and \cref{fig:baseline_comparisson_Nt} we can see that performance increases with the \ac{RIS} size, related to the aforementioned factors. As shown in \cref{fig:baseline_comparisson_Nt}, for both the Linear and Neural equalization methods, increasing the compression factor $\zeta$ beyond a certain threshold no longer yields performance improvements when the \ac{RIS} size is limited to $2\%$ or $6\%$. This plateau highlights the bottleneck imposed by the \ac{RIS}, which restricts the system’s ability to leverage additional complexity. In contrast, this effect is not observed at higher \ac{RIS} sizes, where performance continues to improve with increasing $\zeta$, suggesting that the limiting factor shifts from the \ac{RIS} to model complexity. Similarly, no such bottleneck is observed in \cref{fig:baseline_comparisson_K}, where the \ac{RIS} size scales proportionally with $K$, and thus increases with $\zeta$, as reflected in the figure.

\subsection{Performance with varying SNR}
\begin{figure*}
    \centering
    \includegraphics[width=\figurescaling\linewidth]{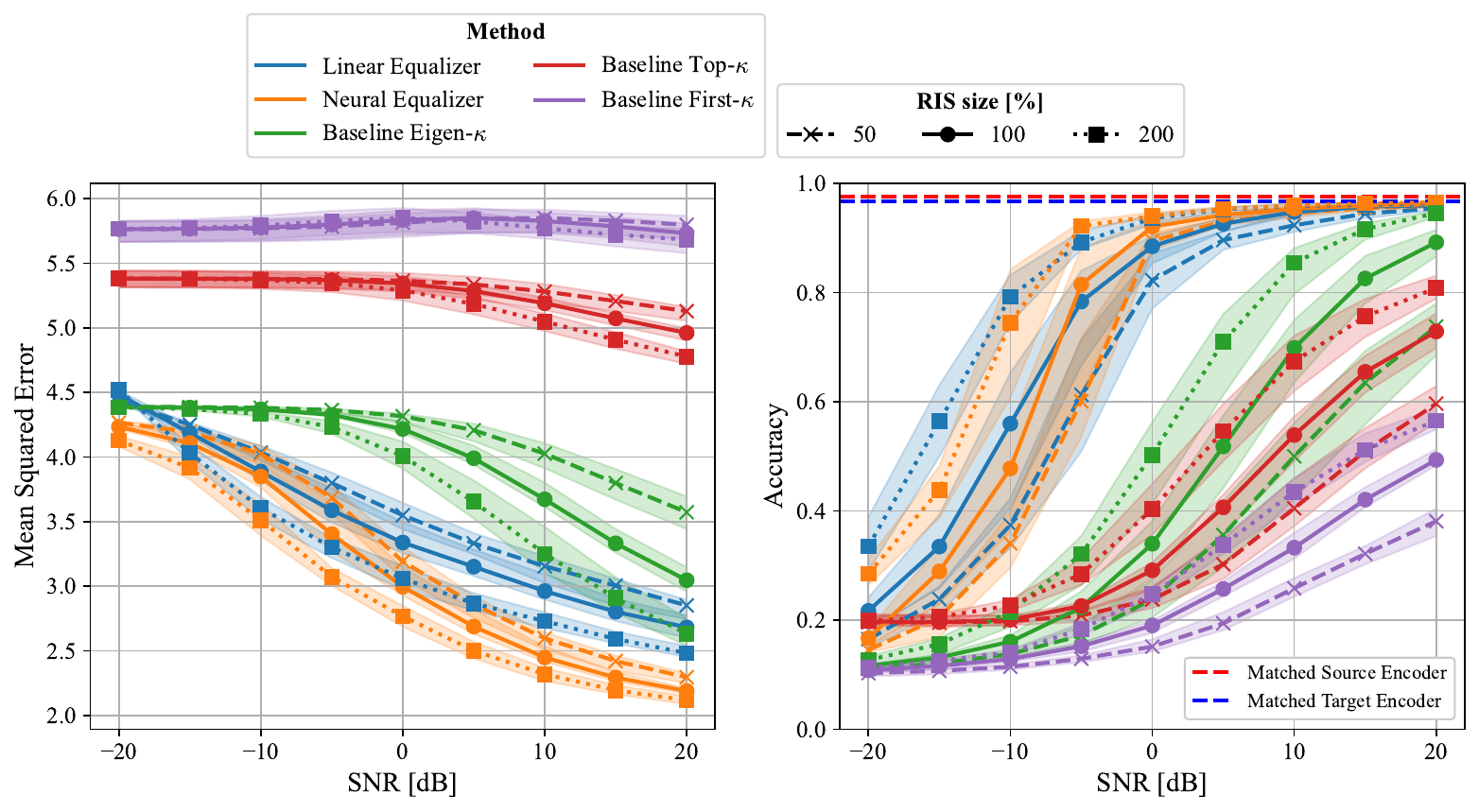}
    \caption{Performance of the proposed methods and baselines with varying \ac{SNR} and different \ac{RIS} sizes for a fixed $\zeta=0.16$ ($N_t=N_r=8$ and $K=8$), $\alpha_d=10^6$ and $\alpha_1=\alpha_2=10$.}
    \label{fig:performance_vs_snr}
\end{figure*}

In \cref{fig:performance_vs_snr} we show the performance of the proposed methods with a varying \ac{SNR} (as defined in \cref{eq:sim_snr}) for $\zeta=0.16$ ($N_t=8$ and $K=8$). The proposed methods outperform the baselines, especially for an $\text{SNR}=0$ dB, where the Linear and Neural Equalizer achieve over 90\% accuracy for the maximum \ac{RIS} size, while the best baseline falls around the 50\% mark. The same is observed in terms of \ac{MSE}, where all baselines are outperformed by the Linear and Neural equalizers. In particular, the Top-$\kappa$ and First-$\kappa$ perform very poorly while the Eigen-$\kappa$ is the best, achieving performance close to the linear equalizer and even presenting a lower \ac{MSE} for an \ac{SNR} of $-20$ dB.

The previously observed discrepancy between \ac{MSE} and Accuracy is observed in these experiments. Notably, while outperformed by the Neural Equalizer in terms of \ac{MSE}, the Linear equalizer achieves better accuracy in the low \ac{SNR} regime (less than $-5$ dB). The same phenomenon can be observed for the Top-$\kappa$ and the Eigen-$\kappa$ baselines. This motivates, again, the exploration of other objective metrics other than the \ac{MSE} that better align with the Accuracy.




\subsection{Effect of training set size $N$}

\begin{figure*}
    \centering
    \includegraphics[width=\figurescaling\linewidth]{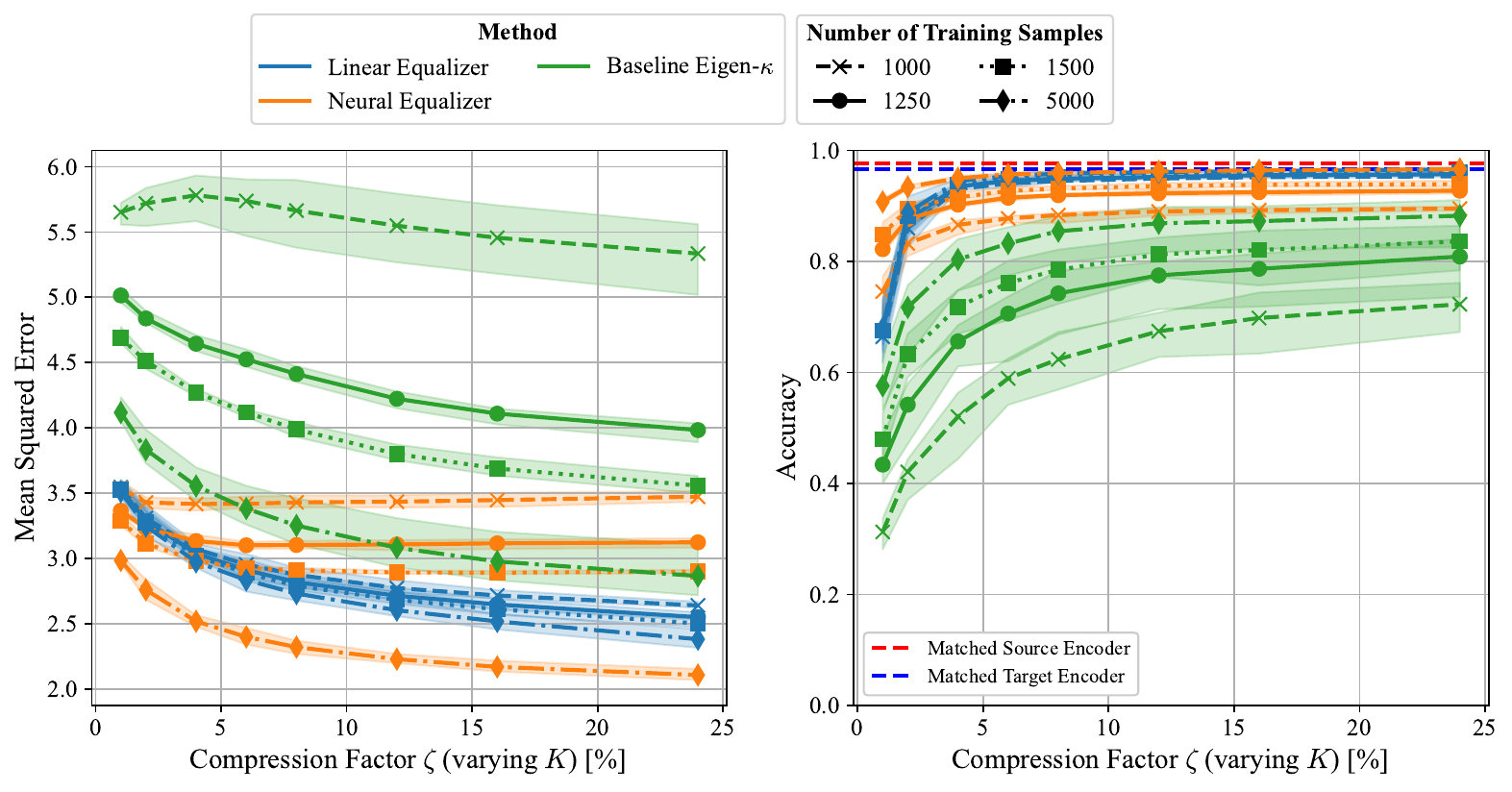}
    \caption{Performance of the proposed methods and the baseline with varying compression factor and different values of $N$. Fixed values of $N_t=N_r=8$, $N_\ris=16$, $\alpha_d=10^6$, $\alpha_1=\alpha_2=10$ and \ac{SNR}$=10$ dB.}
    \label{fig:performance_vs_ntrain}
\end{figure*}

In \cref{fig:performance_vs_ntrain} we show the performance of the proposed methods and the baseline Eigen-$\kappa$ with varying compression factor for different numbers of training samples for $\zeta=0.16$ ($N_t=8$ and $K=9$) and \ac{SNR}$=10$ dB. Both proposed methods are more robust to the number of training samples and perform better than the Eigen-$\kappa$ baseline. Since the proposed linear and neural approaches include noise in the design of the equalizer, they are naturally regularized, making the generalization better. Nevertheless, among the proposed methods, it is natural that the Neural Equalizer is more sensible to the number of training samples, due to its higher complexity which can lead to overfitting. On the other hand, the Linear Equalizer is highly robust to the number of training samples, presenting similar performance across all different values. 

\subsection{Complexity analysis}

While the Neural equalizer consistently outperforms the Linear one across all experiments, it incurs higher computational complexity. To quantify the trade-off between complexity and performance, we evaluate the number of \acp{FLOP} required to process a single sample for both models, based on their respective architectures. A complex matrix-vector multiplication between a matrix of size $a \times b$ and a vector of size $b$ requires $a(8b - 2)$ \acp{FLOP}. When including bias terms, each dense layer in a \ac{DNN} contributes $8ab$ \acp{FLOP}. For a layer with sparsity $s$ (i.e., a fraction $s$ of weights being zero), and assuming the zeros are evenly distributed, the cost reduces to $(1 - s)8ab$ \acp{FLOP}.

\textbf{Linear Model \acp{FLOP}:}
The Linear model uses matrices $\F \in \bC^{KN_t \times \sdimC}$ and $\G \in \bC^{\tdimC \times KN_r}$, yielding:
\begin{equation}\label{eq:linear_flops}
\text{FLOPs}_{\text{linear}} = KN_t\parenthesis{8\sdimC - 2} + \tdimC\parenthesis{8KN_r - 2}
\end{equation}

\textbf{Neural Model \acp{FLOP}:}
The Neural model consists of $\f_\eta$ (input: $\sdimC$, hidden: $\tdimC$, output: $KN_t$) and $\g_\nu$ (input: $KN_r$, hidden: $\tdimC$, output: $\tdimC$), both with \ac{GELU} activations in the hidden layer. The total cost is:
\begin{align}
\text{FLOPs}_{\text{neural}} &= 8(1 - s) \Bigg ( \sdimC \tdimC + \tdimC K N_t \nonumber \\
&\quad + KN_r \tdimC + \tdimC^2 \Bigg )+ 2c \tdimC \label{eq:neural_flops}
\end{align}

where $c \approx 100$ denotes the cost in \acp{FLOP} of computing one \ac{GELU} unit.

\begin{figure*}
    \centering
    \includegraphics[width=\linewidth]{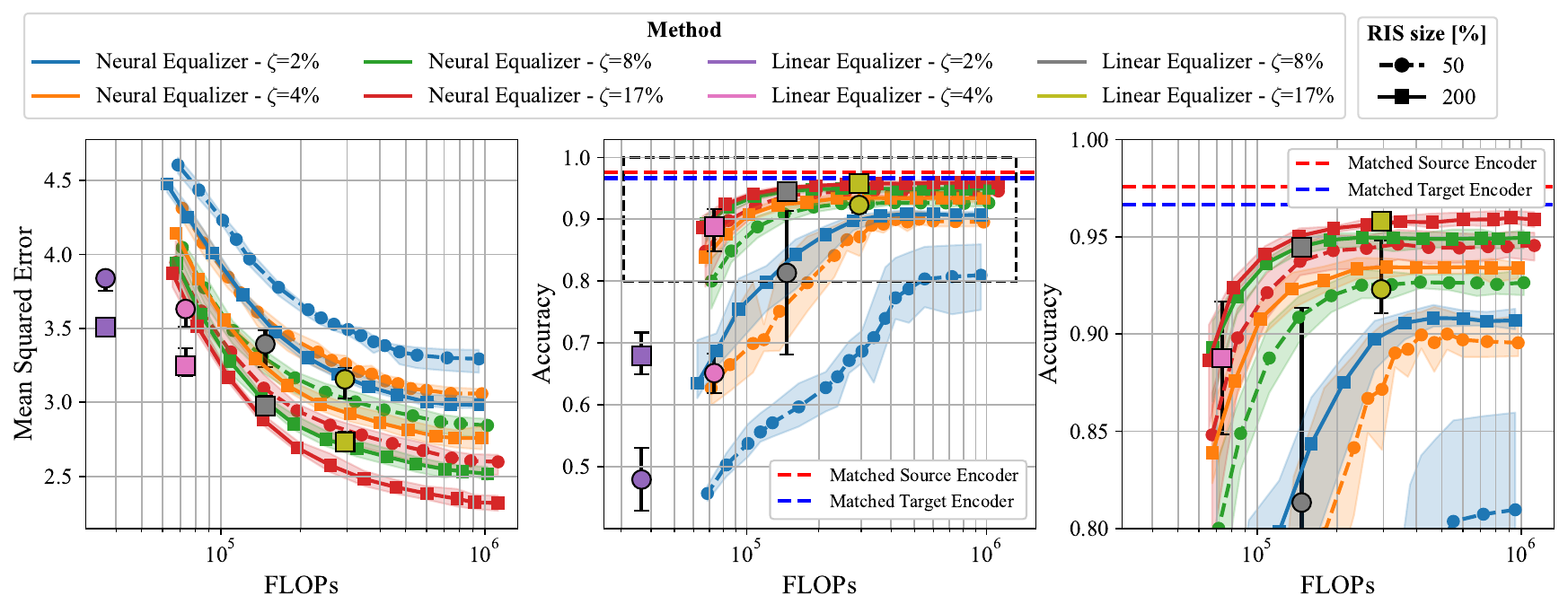}
    \caption{Accuracy and detail as a function of the number of \acp{FLOP} for the Linear and Neural equalizers with varying \ac{RIS} sizes and compression factors. Fixed values of $N_t=N_r=8$, \ac{SNR}$=10$ dB, $\alpha_d=10^6$ and $\alpha_1=\alpha_2=10$. Points for the Linear equalizer are increased in size for better readability.}
    \label{fig:accuracy_vs_flops}
\end{figure*}

In \cref{fig:accuracy_vs_flops}, we plot the \ac{MSE} and accuracy of both models against their corresponding \acp{FLOP}, computed from \cref{eq:linear_flops,eq:neural_flops}, for various \ac{RIS} sizes and compression factors $\eta$ (with fixed $N_t = 8$ and varying $K$) with an \ac{SNR} of 10 dB. Neural model sparsity is controlled via the parameter $\beta$ (see \cref{alg:hard_thresholding_eta_nu}) across the range $\{1, 2, 4, 6, \ldots, 20\}$, achieving up to $s = 98\%$.

As expected, both metrics improve with increasing \acp{FLOP}. For the Linear model, this increase is accompanied by a higher number of transmitted symbols, making it more difficult to isolate the effect of \acp{FLOP} alone. Nevertheless, the Neural model exhibits the same positive trend, confirming that increased computational effort enhances performance. The accuracy of the Neural equalizer remains constant across a wide range of \acp{FLOP}, indicating that the model is indeed over-parameterized and that pruning a significant portion of its weights has a negligible impact on performance. In contrast, the \ac{MSE} improves consistently with increasing \acp{FLOP}, though the gains become less pronounced at higher values. This diminishing return suggests that additional computational complexity beyond a certain threshold offers limited benefit. Finally, the results highlight the importance of the communication bottleneck: for a fixed number of \acp{FLOP}, performance varies significantly with $\zeta$. This indicates that increasing the communication budget may be more effective than increasing computational complexity.

When comparing both models, a sparse Neural equalizer appears to outperform the Linear one in the high-complexity regime, achieving similar or even superior performance. In contrast, the Linear equalizer remains more effective in the low-complexity regime. However, in the scenarios where the Linear equalizer performs better, the overall performance is typically inadequate for the task (e.g., accuracy below 90\%), making it less practical. Additionally, the Neural equalizer is less sensitive to the size of the \ac{RIS}, further supporting its advantage over the Linear model. These results suggest that the Neural equalizer is the preferred choice for addressing semantic mismatch. However, the choice of equalizer should not be based solely on the complexity of processing a single sample but also on the complexity of training. Since the equalization process must adapt to dynamic changes in both the channel and the users, the design complexity of the equalizer is critically important. As demonstrated in previous results, the Linear equalizer performs best with smaller dataset sizes, which directly impacts training complexity and may make it more suitable for certain scenarios. We leave a thorough investigation of the complexity of the training process both for a Linear and a Neural (dense and sparse) equalizer for future work.

\subsection{RIS for channel adaptation}
\begin{figure*}
    \begin{subfigure}[b]{\linewidth}
        \centering
        \includegraphics[width=\figurescaling\linewidth]{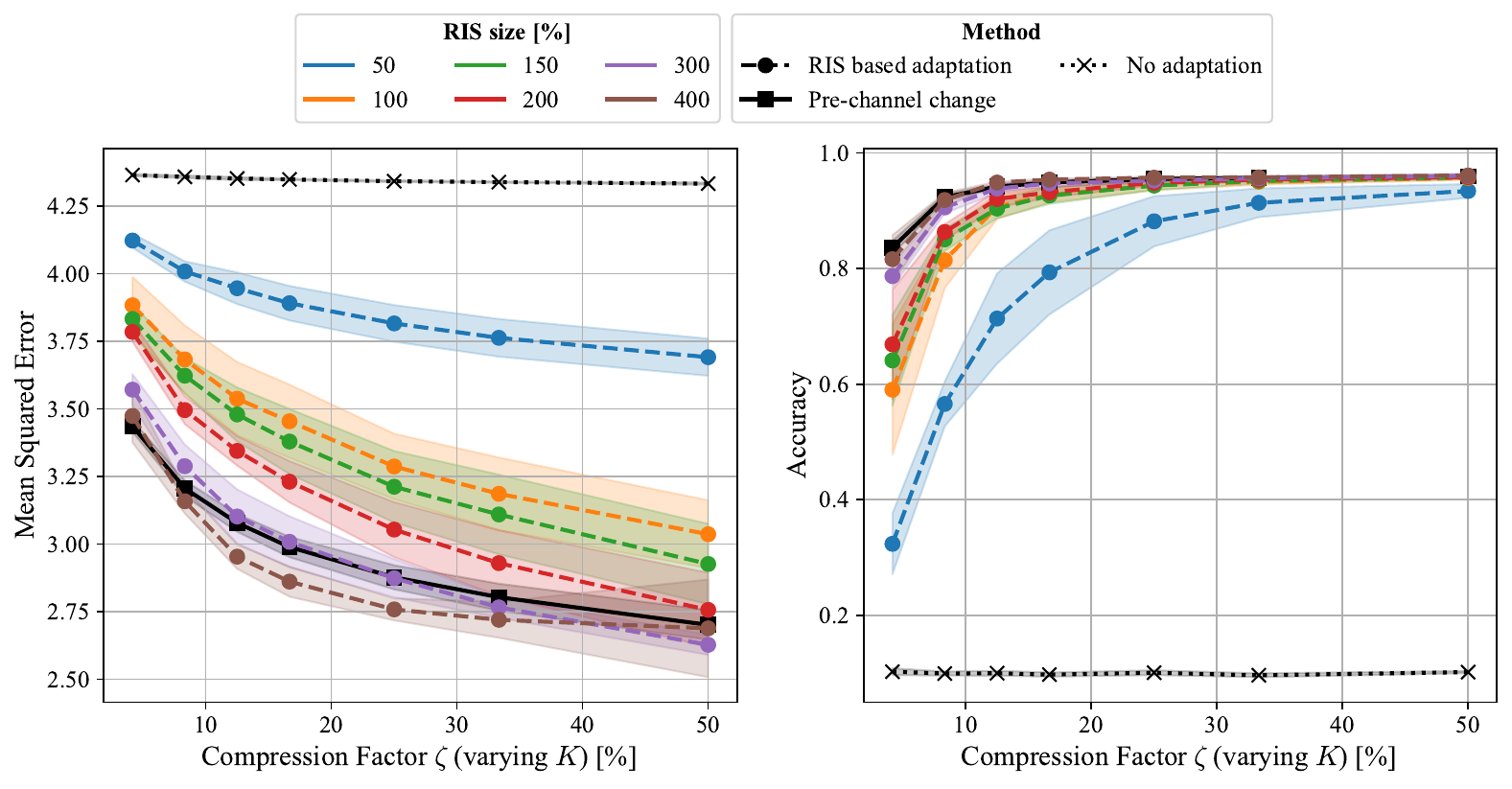}
        \caption{Linear Equalizer.}
        \label{fig:channel_adaptation_linear}
    \end{subfigure}
    \hfill
    \begin{subfigure}[b]{\linewidth}
        \centering
        \includegraphics[width=\figurescaling\linewidth]{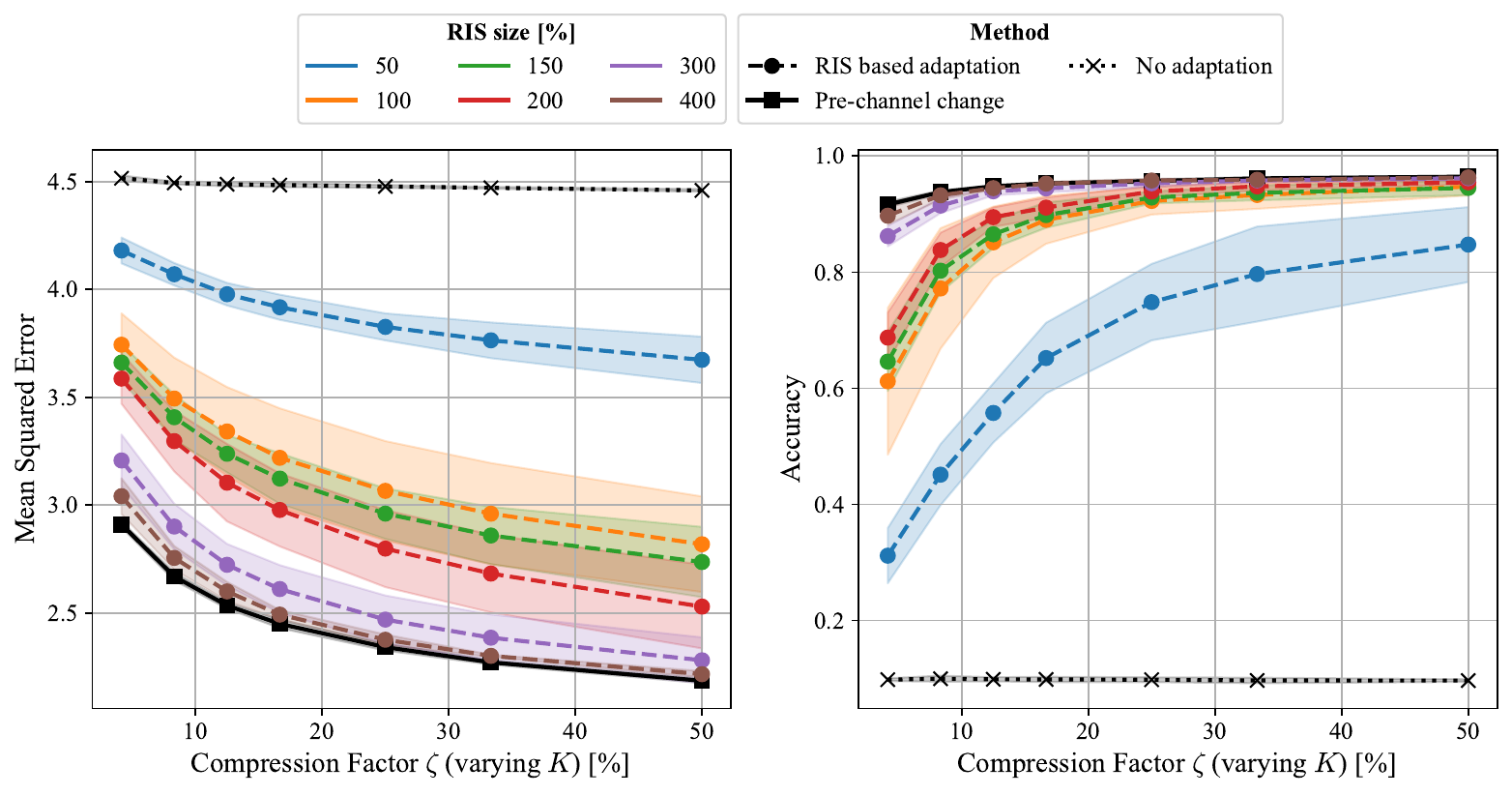}
        \caption{Neural Equalizer}
        \label{fig:channel_adaptation_neural}    
    \end{subfigure}
    \caption{Channel adaptation through a \ac{RIS} module. We set $N_t=N_r=4$ and $\text{SNR}=10$ dB.}
\end{figure*}
In this section, we analyze the potential use of \ac{RIS} to adapt to the communication channel changes. For this, we perform an experiment where we first design the proposed Neural and Linear equalization method in a communication scenario where the direct channel $\H_d$ has a low attenuation $\alpha_d$ and there is no \ac{RIS}, i.e., $\H_e=\H_d$. After this, we simulate a blocking of the direct channel by modifying $\H_d$ and re-sampling its coefficients with higher attenuation $\alpha_d'$, which yields $\H_d'$. With this new blocked channel, we introduce a \ac{RIS} parallel channel ($\H_e^\ris=\tilde{\H}_d'+\H_\ris$) and we optimize the \ac{RIS} parameters to minimize the \ac{MSE} of recovered samples while leaving the equalizer fixed. In \cref{fig:channel_adaptation_linear} and \cref{fig:channel_adaptation_neural} we show the performance of the proposed methods before the channel blocking, after the channel blocking without \ac{RIS} or any type of adaptation, and the \ac{RIS} based adaptation for a Linear and \ac{DNN}-based equalizer, respectively. We set $\alpha_d=10$, $\alpha_d'=10^6$, $\alpha_1=\alpha_2=10$, $N_t=8$ and an \ac{SNR} of 10 dB. To optimize the \ac{RIS} we perform the proposed \ac{PGD} step introduced in \cref{eq:ris_step}, leveraging the same dataset previously used to train the Linear and Neural equalizers. We perform only 10 \ac{PGD} steps to limit the complexity of the equalization process.


As expected, with no adaptation, the performance of the system completely degrades, with accuracy falling to 10\%, i.e., random decision. On the other hand, the \ac{RIS}-based adaptation is able to compensate for the modification of the channel change. As the number of \ac{RIS} elements $N_\ris$ increases, the adaptation is better, leading to reduced \ac{MSE} and increased Accuracy. In \cref{fig:channel_adaptation_linear}, for the highest compression factor, the performance for a \ac{RIS} size of $300\%$ is better than that of $400\%$, which can be explained by the fact that, as the number of \ac{RIS} elements increases, the convergence requires a higher number of iterations. For $400\%$, the limited number of iterations may have been insufficient for the parameters to fully converge.

For the Linear Equalizer, when the \ac{RIS} size is 3 or more times the size of the \ac{MIMO} channel, the performance of the adapted system surpasses that of the pre-channel change. This can be explained by the fact that each \ac{RIS} element introduces a new optimization variable, which allows for a higher complexity equalizer. This way, the \ac{RIS} module not only adapts the channel to the one used to obtain $\F$ and $\G$ but actually enhances the semantic equalization capabilities.

For the Neural equalizer, this phenomenon is not observed as performance is limited by the pre-channel change scenario. This can be explained by the fact that a Neural Equalizer is more complex and therefore benefits less from the inclusion of \ac{RIS} parameters. As the number $N_\ris$ increases, the \ac{RIS} can compensate for the physical channel change but not further improve the semantic equalization, which explains why the performance never surpasses that of the Pre-channel change model. These results highlight that \ac{RIS} can not only work as a channel equalizer but also enhance the performance of a semantic equalizer.

\section{Conclusions}\label{sec:conclusions}

In this work, we proposed a novel methodological framework for semantic equalization that jointly performs semantic alignment, data compression, and channel equalization, distributed across the transmitter, receiver, and channel via a \ac{RIS} module. We formulated the equalization task as a constrained \ac{MMSE} optimization problem, where the goal is to jointly determine the transmitter-side (pre), receiver-side (post) equalizers, and the \ac{RIS} phase configuration to minimize the semantic distortion between the transmitted and recovered symbols. Adopting a data-driven methodology, we leveraged a dataset of semantic pilots and developed two strategies to solve the problem. The first is a linear approach, where both pre and post equalizers are linear models, and the optimization is solved using an \ac{ADMM} algorithm. The second solution is based on \acp{DNN}, trained via conventional gradient-based \ac{ML} techniques. Experimental results demonstrate that both proposed methods significantly outperform disjoint semantic and physical channel equalization schemes under various compression levels, \ac{RIS} sizes, \ac{SNR} conditions, and training dataset sizes. We performed a complexity assessment of the proposed methods and showed that a Neural Equalizer trained with sparsity constraints requires a similar number of \acp{FLOP} per sample to achieve comparable performance as the Linear model, while the Linear model might benefit from lower training complexity due to its effectiveness on smaller datasets. Furthermore, we investigated the adaptability of the \ac{RIS} in response to channel variations. Our results reveal that, beyond its classical role in channel adaptation, the \ac{RIS} can enhance semantic alignment by embedding computational capabilities within the channel, thus contributing directly to the end-to-end semantic communication process. While the chosen optimization metric in this work is the \ac{MSE} between the transmitter and receiver latent spaces, our results indicate that it may be suboptimal due to its relatively weak correlation with task-specific performance, such as classification accuracy. This observation, also reported in \cite{huttebraucker2024relative}, highlights the need for performance-oriented alignment metrics that better reflect end-task objectives. Inspired by the results obtained with the proposed baselines, we argue that a weighted version of the \ac{MSE}—where errors in larger components are prioritized over smaller ones—could serve as a promising alternative to the standard \ac{MSE}. This approach has the advantage of being easily integrated into our algorithm without requiring any additional information about the receiver. In contrast, task-specific metrics like classification cross-entropy depend on detailed knowledge of the receiver's output space, which may not always be available or practical to obtain.

Our proposed framework  lays a solid foundation for design, deploying and operating practical Multi-User \ac{SemComs} in real-world scenarios by addressing the semantic misalignment problem while incorporating the physical characteristics of the communication channel. Although this work constitutes a significant first step, several research directions remain open to fully realize the potential of \ac{SemComs}. First, in this study, we employed generic \ac{ML} models that are communication-agnostic. A natural next step is to integrate our semantic equalization framework with dedicated, communication-aware \ac{DJSCC} models, which are optimized for transmission tasks. While our proposed methods can work without any further adaptation in this case, the communication aware nature of \ac{DJSCC} protocols could be leveraged to reduce the complexity of the proposed solutions. Such integration is essential to evaluate the practicality and performance of the system in real-world deployments. 
Finally, a deeper investigation into the role of \ac{RIS} as an equalization mechanism, particularly in comparison to traditional equalization strategies is needed. Understanding the full potential of programmable smart wireless environments in semantic alignment and joint communication-computation tasks may unlock new capabilities for next-generation wireless systems \cite{strinati2021reconfigurable}.

\section*{Acknowledgments}
The present work was supported by the EU Horizon 2020 Marie Skłodowska-Curie ITN Greenedge (GA No. 953775), by the Horizon SNS-JU Project ``6G-GOALS" (GA No. 101139232), and by the French project funded by the program \quotes{PEPR Networks of the Future} of France 2030.



\bibliographystyle{ieeetr}
\bibliography{include/bib}

\end{document}